\documentclass[11pt]{article}

\usepackage[preprint]{acl}

\usepackage{times}
\usepackage{latexsym}

\usepackage[T1]{fontenc}

\usepackage[utf8]{inputenc}

\usepackage{microtype}

\usepackage{inconsolata}

\usepackage{algorithm}
\usepackage{algpseudocode}

\usepackage{adjustbox}
\usepackage{amsthm}
\usepackage{amsmath}
\usepackage{amssymb}

\usepackage[utf8]{inputenc} 
\usepackage[T1]{fontenc}    
\usepackage{url}            
\usepackage{booktabs}       
\usepackage{tabularx}
\usepackage{amsfonts}       
\usepackage{nicefrac}       
\usepackage{microtype}      
\usepackage{lipsum}         
\usepackage{enumitem}
\usepackage{graphicx}
\usepackage{natbib}
\usepackage{doi}
\usepackage{epigraph}
\usepackage{multirow}
\usepackage{placeins}
\usepackage[table]{xcolor}

\title{LaRe: Latent Refocusing for Multimodal Reasoning}

\author{Jizheng Ma$^{1,2}$, Xiaofei Zhou$^{1,2\ddagger}$, Geyuan Zhang$^{1,2}$, Yanlong Song$^{1,2}$, Han Yan$^{1,2}$ \\
        $^{1}$Institute of Information Engineering, Chinese Academy of Sciences \\ 
        $^{2}$School of Cyber Security, University of Chinese Academy of Sciences \\
        \small{\texttt{zhouxiaofei@iie.ac.cn}}}

\begin{document}
\maketitle
\begin{abstract}
Chain of Thought (CoT) reasoning enhances logical performance by decomposing complex tasks, yet its multimodal extension faces a trade-off. The prevailing Thinking with Images paradigm achieves visual refocusing by explicitly cropping image regions, yet incurs rapidly growing computational overhead. The emerging line of latent-space reasoning reduces token consumption, but lacks the capacity for dynamic refocusing. We argue that this trade-off stems from a tacitly accepted premise that effective visual refocusing must occur in the form of explicit tokens. Building on this, we propose \underline{La}tent \underline{Re}focusing (LaRe), a new multimodal reasoning paradigm in which visual refocusing takes place entirely within the latent space. We further design a semantic augmentation training strategy that ensures the semantic structure of the latent space through visual reconstruction objective. Experimental evaluations demonstrate that LaRe improves average accuracy by 7.6\% compared to existing baselines while reducing the number of tokens required for inference by 59.7\%. When scaled to a 8B-parameter Vision-Language Model backbone, LaRe achieves performance comparable to state-of-the-art methods, demonstrating the efficacy of our proposed latent refocusing paradigm for multimodal reasoning.
\end{abstract}

\section{Introduction}

The Chain of Thought (CoT) reasoning mechanism enhances the reasoning capabilities of large language models (LLMs) by decomposing complex problems through explicit intermediate steps \citep{wei2022chain,zhou2024miceval,yao2023tree,openai2024openaio1card,zhang2024chain,wang2022self,brown2020language,minaee2024large}. With the rapid development of Multimodal Large Language Models (MLLMs) in recent years, extending Chain of Thought to multimodal scenarios has become increasingly important \citep{xu2023multimodal,li2025surveystateartlarge}. This trend has given rise to multimodal reasoning as an emerging research direction that is gradually becoming an important technique for achieving artificial general intelligence (AGI) \citep{wang2025multimodal}.

\begin{figure}[t]
    \centering
    \includegraphics[width=1.0\linewidth]{motivation.pdf}
    \caption{Illustrative comparison across Thinking with Images, Latent Space Reasoning, and our proposed Latent Refocusing (LaRe) framework. LaRe performs refocusing within latent space, achieving more flexible reasoning.}
    \label{fig:figure1_motivation}
\end{figure}

Currently, the predominant paradigm for multimodal reasoning is Thinking with Images (illustrated in Figure \ref{fig:figure1_motivation} (a)), which allows the model to explicitly crop and rescale image regions during inference so that the reasoning chain can refocus on visual evidence \citep{fu2025refocus,gao2025interleaved}. Despite its effectiveness, this paradigm requires every visual refocus to be injected into the reasoning chain in the form of explicit visual tokens, causing the number of reasoning tokens to grow rapidly with the number of refocusing \citep{wang2025multimodal}. In scenarios involving complex visual evidence, such overhead quickly becomes prohibitive.

To address this issue, a promising emerging direction is Latent Space Reasoning (illustrated in Figure \ref{fig:figure1_motivation} (b)) \citep{li2025implicit,hao2024training}, which shifts the reasoning process from explicit tokens to latent representations. By performing reasoning within the latent space, such methods substantially reduce token consumption. However, current explorations of latent space reasoning remain at an early stage and lack the dynamic refocusing ability found in paradigms such as Thinking with Images. Consequently, the reasoning process cannot anchor itself to key visual information. Most existing methods either compress reasoning into a static sequence \citep{su2025token,xu2025softcot} or treat latent visual features as static initialization inputs \citep{li2025latent}. Moreover, prevailing training objectives are largely confined to conventional autoregressive language modeling leaving the latent space without effective supervisory signals and rendering latent representations a black box.

These two lines of work reveal a fundamental tension in multimodal reasoning: under existing paradigms, the ability to refocus visual evidence and the token efficiency of the reasoning process cannot be achieved simultaneously. Thinking with Images trades token overhead for visual refocusing, whereas latent reasoning trades the capacity for visual refocusing for efficiency. We argue that this tension is not intrinsic, but stems from a tacitly accepted premise that effective visual refocusing must occur in the form of explicit tokens. Once this premise is abandoned, visual refocusing can take place within a continuous latent space, preserving the flexibility of multi-round refocusing while avoiding token inflation. Motivated by this observation, we propose \underline{La}tent \underline{Re}focusing (LaRe), a novel multimodal reasoning paradigm that realizes visual refocusing entirely within the latent space as depicted in Figure \ref{fig:figure1_motivation} (c).

Specifically, latent refocusing does not rely on explicit image tokens for refocusing. Instead it introduces a set of iteratively evolving latent representations. At each iteration a lightweight module refocuses on visual information conditioned on the current reasoning state and fuses it into the reasoning state to form latent tokens. These tokens then guide the next reasoning step producing an iterative refocusing process. The latent tokens inhabit a latent space that encodes the evolving intermediate reasoning states. To shape this space into a semantically meaningful and interpretable structure, we further propose a semantic augmentation training strategy based on visual reconstruction. Inspired by the ability of denoising processes to encourage models to learn the underlying data manifold \citep{karras2022elucidating,chen2023score}, this strategy employs a denoising objective for visual reconstruction, thereby guiding the latent space toward a structured semantic layout that is conducive to reasoning. The reconstruction objective ensures that the latent space retains the fine grained visual information required for reasoning. 
The contributions of this work are as follows:

\begin{itemize}[left=0pt]
\item We propose Latent Refocusing, which enables the model to dynamically anchor critical visual evidence within a continuous latent space, thereby achieving more flexible and efficient inference. To the best of our knowledge, LaRe is among the first to investigate visual refocusing as an explicit mechanism within latent space for multimodal reasoning.

\item We propose a training strategy tailored for latent-space reasoning. By leveraging visual reconstruction with a denoising objective, we guide the model to produce semantic representations in the latent space that are beneficial for reasoning, thereby enhancing its ability to capture complex spatial layouts and implicit visual relationships.

\item Across multiple benchmarks, LaRe improves average accuracy by 7.6\% over strong baselines while reducing inference token consumption by 59.7\%. Qualitative analysis further shows that the latent representations are semantically coherent and interpretable, offering new directions for future research on latent space reasoning.

\end{itemize}

\begin{figure*}[ht]
    \centering
    \includegraphics[width=1.0\linewidth]{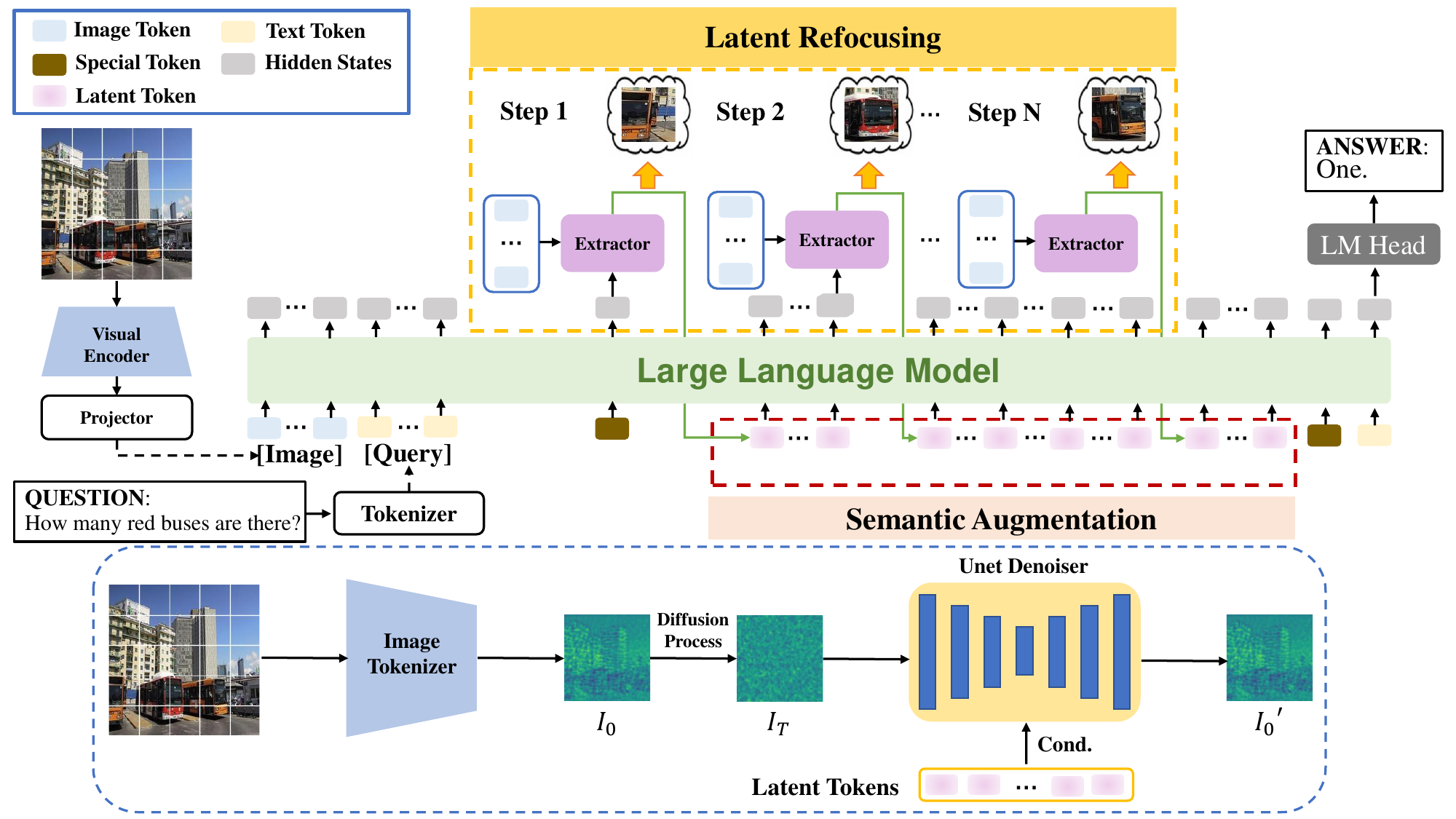}
    \caption{Overview of our proposed LaRe paradigm. The framework comprises Latent Refocusing (top), where the LLM and Extractor generate latent tokens by adaptively refocusing on visual evidence conditioned on reasoning states, and Semantic Augmentation (bottom), which leverages a diffusion-based visual reconstruction task to provide supervision, ensuring that latent tokens maintain both high-level semantics and fine-grained visual fidelity.}
    \label{fig:figure2_overview}
\end{figure*}

\section{Related Work}

\paragraph{Multimodal Reasoning.} Chain of Thought (CoT) reasoning was originally designed to augment the capabilities of Large Language Models (LLMs) through explicit reasoning steps \citep{wei2022chain} and has since been expanded into the multimodal domain \citep{wang2025multimodal}. Early methodologies primarily relied on textual rationales \citep{zhang2023multimodal}, though such approaches frequently struggle to capture fine grained spatial details. To bridge this gap, the Thinking with Images paradigm incorporates dynamic visual information into the reasoning process. For example, ReFocus \citep{fu2025refocus} guides attention by highlighting structured regions, while Visual CoT \citep{rose2023visual} inserts image states into intermediate reasoning steps. Other methodologies such as Chain of Image \citep{meng2023chain} and VisualSketchpad \citep{hu2024visual} utilize auxiliary sketches for inference, whereas ICoT \citep{gao2025interleaved} constructs interleaved multimodal sequences. While these methods improve reasoning accuracy, they incur additional computational overhead.

\paragraph{Latent Space Reasoning.} Recent research trends have increasingly shifted toward latent space reasoning, which processes intermediate reasoning steps within the internal states of a model without explicit verbalization \citep{zhu2025surveylatentreasoning, li2025implicit}. Existing works explore latent optimization at the token level \citep{tack2025llm, sun2025enhancing, gong2025latent} or the trajectory level \citep{hao2024training, shen2025efficient, shen2025codi, zhang2025soft, kong2025scalable} while incorporating signal guided control mechanisms \citep{herel2024thinking, li2025seek} and recurrent execution structures across layers \citep{saunshi2025reasoning, mohtashami2023cotformer}. Nevertheless, the majority of these approaches are restricted to the textual manifold. Recent multimodal latent reasoning methods, including MCOUT \citep{pham2025multimodal}, Mirage \citep{yang2025machine}, and LVR \citep{li2025latent}, often lack an effective refocusing mechanism. This deficiency results in weak visual anchoring during reasoning, which limits their overall expressiveness and scalability.

\section{Methodology} \label{sec:LaRe}
In this section, we present LaRe, a novel multimodal reasoning paradigm that performs refocusing within a shared latent space. LaRe consists of two core components: (1) an iterative Latent Refocusing mechanism that targets reasoning-relevant visual information through the introduction of dynamically evolving latent representations; and (2) a semantic augmentation training strategy designed to represent multimodal information within the latent space to facilitate the latent refocusing process.

\subsection{Latent Refocusing}
LaRe endows the model with flexible multimodal reasoning capabilities through an iterative refocusing mechanism as illustrated in Figure~\ref{fig:figure2_overview}. We model the reasoning process as a set of latent representations that evolve within the latent space. These representations encapsulate reasoning information and concentrate on the most relevant visual data.

Specifically, at each iteration step, we adaptively focus on and fuse reasoning-relevant visual information based on latent textual information to obtain latent tokens. Formally, the process of the $k$-th iteration step is expressed as follows:

\begin{align}
{H}_T^{(k)} &= {H}_{\text{LLM}}^{(k)}([{V}_T; {T}; {Z}^{(1:k-1)}]), \\
{Z}^{(k)} &= \text{Extractor}({H}_T^{(k)}, {V}_T).
\end{align}

In these equations, ${V}_T \in {\mathbb{R}}^{m\times d}$ denotes the visual tokens obtained from the original image via a visual encoder and multimodal projector \citep{bai2023qwen,li2023blip}, $T \in {\mathbb{R}}^{n\times d}$ represents the text embeddings of the input query, ${Z}^{(k)} \in {\mathbb{R}}^{l\times d}$ represents the latent tokens generated at the current step, and ${Z}^{(1:k-1)}$ constitutes the historical sequence of latent tokens generated in previous iterations. In this context, $m$, $n$, and $l$ denote the sequence lengths of visual tokens, textual tokens, and latent tokens, respectively, while $d$ represents the hidden dimension of the LLM.

During this process, the Large Language Model (LLM) functions as the decision hub. The LLM processes the concatenated input sequence and extracts hidden states, denoted as ${H}_{\text{LLM}}^{(k)}$, to generate intermediate hidden representations ${H}_T^{(k)} \in {\mathbb{R}}^{l\times d}$ from every layer. These representations encapsulate the textual reasoning information of the model at the current step as described by \citep{hao2024training}. To exploit the representational characteristics inherent across all layers of the LLM \citep{skean2025layer}, we set the length $l$ to match the number of layers in the LLM. Subsequently, the Extractor, acting as a lightweight semantic aggregation module, focuses on reasoning-relevant visual information from ${V}_T$ based on ${H}_T^{(k)}$ and fuses it into latent tokens. The resultant latent tokens $Z^{(k)}$ encapsulate critical reasoning evidence and drive subsequent reasoning steps.

\paragraph{Extractor.}
The Extractor is a lightweight Transformer module designed to focus on task-relevant evidence from the visual features ${V}_T$ through an $N$-layer process. Each layer performs internal logic modeling followed by cross-modal refocusing, both stabilized by residual connections to preserve reasoning context. Starting with ${L}^{(0)} = {H}_T^{(k)}\in {\mathbb{R}}^{l\times d}$, the update rules for the $i$-th layer and the final synthesis of latent tokens ${Z}^{(k)}$ are:

{\small
\begin{align}
{L}_{\text{SA}}^{(i)} &= \text{Self-Attn}\big(\text{Norm}(\text{Linear}({L}^{(i-1)}))\big) + {L}^{(i-1)}, \\
{L}_{\text{CA}}^{(i)} &= \text{Cross-Attn}\big(\text{Norm}({L}_{\text{SA}}^{(i)}), {V}_T\big) + {L}_{\text{SA}}^{(i)}, \\
{Z}^{(k)} &= \text{FFN}\big(\text{GLU}({L}_{\text{CA}}^{(N)})\big).
\end{align}}

By employing the latent state as a query to probe ${V}_T$, the Extractor extracts the precise visual signals required for the current reasoning step. The resulting ${Z}^{(k)}$ serves as a semantic carrier to drive the subsequent iteration of the large language model.

\paragraph{Inference Phase.}
After reaching the preset number of iterations \(K\), we prepend and append the \texttt{$<\vert\texttt{bot}\vert>$} and \texttt{$<\vert\texttt{eot}\vert>$} special tokens to the latent sequence to delimit the boundaries of the continuous iteration, attach this sequence to the original image-text input, and then let the model generate text tokens autoregressively.

\subsection{Semantic Augmentation}
To prevent the loss of critical low-level details during latent compression and refocusing, we propose a semantic augmentation strategy centered on visual reconstruction. This objective requires the generated latent tokens to reconstruct fine-grained visual representations, which ensures that the latent representations maintain high-level semantics while preserving necessary evidential fidelity.

Inspired by diffusion models and denoising autoencoders, we adopt a denoising training scheme rather than simply reconstructing static features through latent tokens. This objective addresses significant spatial redundancy \citep{he2022masked} and provides meaningful supervision signals centered on visual reconstruction. Denoising outperforms standard regression because the introduction of noise serves as implicit data augmentation and regularization, which compels the model to focus on the underlying data manifold \citep{karras2022elucidating,chen2023score} instead of memorizing specific instance values.

The reconstruction objective $\mathcal{L}_{\text{rec}}$ is formalized using a denoising score matching loss:

{\small
\begin{equation}
\mathcal{L}_{\text{rec}} = \mathbb{E} \left[ | \epsilon - \epsilon_\theta( \sqrt{\bar{\alpha}_t} {I}_0 + \sqrt{1-\bar{\alpha}_t} \epsilon, t, {Z} ) |_2^2 \right],
\end{equation}}

where $t \sim \mathcal{U}(0, 1)$ is the sampled timestep, ${I}_0$ denotes the visual representations encoded by the Image Tokenizer, $\epsilon \sim \mathcal{N}(0, {I})$ represents the injected target noise, $\bar{\alpha}_t$ is the predefined noise schedule coefficient, and $\epsilon_\theta$ is a lightweight denoising U-Net \citep{ronneberger2015u} conditioned on all the latent tokens ${Z}$.

The complete training objective combines these losses with the standard next-token prediction loss:

\begin{gather}
\mathcal{L}_{ntp} = -\sum_{i=1}^{|{Y}|} \log P(y_i | {V}_T, {T}, {Z}, y_{<i}), \\
\begin{split}
\mathcal{L}_{total} = \mathcal{L}_{ntp} &+ \lambda_{rec}\mathcal{L}_{rec},
\end{split}
\end{gather}

where $\lambda_{rec}$ is hyperparameter.

\section{Experiments}\label{sec:experiments}

\begin{table*}[htbp]
    \centering
    
    \renewcommand{\arraystretch}{1.2} 
    \setlength{\tabcolsep}{2.5pt} 
    \small 
    
    \caption{Comparison to various multimodal reasoning baselines. The best results are \textbf{bolded}. $^*$Unlike LaRe and the other baseline methods, the post training of CoF and LVR follows their original settings.}
    \label{tab:main}
    
    \begin{adjustbox}{width=1.0\textwidth}
    \begin{tabular}{@{} l 
      c c  
      c c  
      c c  
      c c  
      c c 
      c c
      c c
      @{}}
    \toprule
    \multirow{2}{*}{Method}
      & \multicolumn{2}{c}{MMBench}   & \multicolumn{2}{c}{MMStar}
      & \multicolumn{2}{c}{MMVP}   & \multicolumn{2}{c}{ScienceQA}
      & \multicolumn{2}{c}{POPE} & \multicolumn{2}{c}{${V}^{*}$}
      & \multicolumn{2}{c}{MMMU-Pro} \\
    \cmidrule(lr){2-3} \cmidrule(lr){4-5} \cmidrule(lr){6-7} \cmidrule(lr){8-9} \cmidrule(lr){10-11} \cmidrule(lr){12-13}
    \cmidrule(lr){14-15} 
    & Acc. & \# Tokens & Acc. & \# Tokens & Acc. & \# Tokens & Acc. & \# Tokens & Acc. & \# Tokens & Acc. & \# Tokens & Acc. & \# Tokens \\
    \midrule

    \multicolumn{15}{c}{\textit{Base VLM: Qwen3-VL-2B-Instruct}}\\
    \hline

    Standard CoT                         & 71.4   & 167.4 & 56.6    & 192.1 & 18.4   & 126.5 & 74.4   & 182.5 & 58.2   & 104.2 & 75.0   & 181.2 & 37.0   & 224.2 \\
    MM-CoT \citep{zhang2023multimodal}   & 72.7   & 158.3 & 57.8    & 120.2 & 24.0   & 107.7 & 80.1   & 195.7 & 51.2   & 100.9 & 66.5   & 192.3 & 39.8   & 198.8 \\
    CCoT \citep{mitra2024compositional} & 69.9   & 115.7 & 56.7    & 174.4 & 37.5   & 128.6 & 79.9   & 200.9 & 63.3   & 110.1 & 75.2   & 221.4 & 35.5   & 203.5 \\
    ICoT \citep{gao2025interleaved}      & 71.5   & 133.4 & 53.3    & 164.1 & 40.3   & 108.7 & 77.0   & 174.2 & 65.9   & 104.7 & 76.3   & 199.2 & 40.8   & 214.7 \\
    CoF$^*$ \citep{zhang2025chain}       & \textbf{73.2}   & 162.6 & 58.1    & 208.9 & 50.1   & 137.3 & 72.5   & 195.7 & 65.7   & 128.9 & 77.3   & 255.9 & 41.2   & 274.0 \\
    MCOUT \citep{pham2025multimodal}     & 62.6   & 65.3 & 58.2    & 95.4 & 37.3   & 67.9 & 76.8   & 104.3 & 64.0   & 37.2 & 76.5   & 99.2 & 39.0   & 120.1 \\
    LVR$^*$ \citep{li2025latent}         & 67.5   & 75.0 & 56.6    & 108.2 & 43.9   & 72.4 & 79.2   & 93.8 & 64.5   & 29.5 & 77.4   & 112.7 & 38.9   & 134.7 \\
    \textbf{LaRe (Ours)}               & 71.9   & \textbf{34.6} & \textbf{61.2}    & \textbf{70.5} & \textbf{53.3}   & \textbf{27.2} & \textbf{82.5}   & \textbf{55.0} & \textbf{68.0}   & \textbf{18.9} & \textbf{81.3}   & \textbf{87.2} & \textbf{44.2}   & \textbf{102.4} \\
    \midrule

    \multicolumn{15}{c}{\textit{Base VLM: Qwen3-VL-4B-Instruct}}\\
    \hline

    Standard CoT                         & 81.7   & 165.2 & 65.3    & 217.0 & 23.7   & 146.2 & 79.5   & 186.2 & 55.1   & 99.5 & 80.0   & 183.4 & 52.4   & 226.8 \\
    MM-CoT \citep{zhang2023multimodal}   & 75.9   & 170.6 & 67.8    & 210.9 & 27.5   & 113.0 & 82.7   & 203.1 & 55.7   & 102.7 & 74.5   & 187.2 & 49.2   & 204.2 \\
    CCoT \citep{mitra2024compositional} & 80.5   & 125.4 & 64.2    & 230.6 & 39.3   & 140.9 & 83.4   & 240.2 & 69.1   & 112.8 & 81.5   & 203.4 & 55.5   & 234.9 \\
    ICoT \citep{gao2025interleaved}      & 76.9   & 162.5 & 62.7    & 200.7 & 46.0   & 122.7 & 81.7   & 172.6 & 66.3   & 103.7 & 80.4   & 195.0 & 52.3   & 211.7 \\
    CoF$^*$ \citep{zhang2025chain}       & 78.6   & 190.8 & 70.6    & 224.4 & 51.8   & 145.8 & 80.6   & 205.5 & 71.0   & 164.1 & 83.7   & 271.1 & 53.7   & 269.1 \\
    MCOUT \citep{pham2025multimodal}     & 73.8   & 72.3 & 61.4    & 145.5 & 45.7   & 69.2 & 78.8   & 97.5 & 64.5   & 56.0 & 82.5   & 104.8 & 52.0   & 124.5 \\
    LVR$^*$ \citep{li2025latent}         & 79.0   & 80.7 & 68.5    & 125.3 & 56.2   & 72.9 & 80.3   & 101.4 & 69.4   & 42.4 & 83.7   & 110.9 & 53.9   & 144.4 \\
    
    \textbf{LaRe (Ours)}               & \textbf{82.1}   & \textbf{49.0} & \textbf{72.8}    & \textbf{82.7} & \textbf{57.6}   & \textbf{32.3} & \textbf{87.5}   & \textbf{65.2} & \textbf{77.4}   & \textbf{29.0} & \textbf{87.0}   & \textbf{93.5} & \textbf{58.4}   & \textbf{103.5} \\
     
    \bottomrule
    \end{tabular}
    \end{adjustbox}
\end{table*}

\subsection{Datasets and Evaluation}
To assess the performance of the proposed LaRe framework, we select seven benchmarks covering general multimodal understanding, fine-grained perception, hallucination detection, and complex reasoning. For general reasoning, we utilize MMBench \citep{liu2024mmbench} for comprehensive multimodal evaluation and ScienceQA \citep{lu2022scienceqa} for interdisciplinary scientific reasoning, as these tasks require stable logical chains and effective vision-language alignment. To evaluate the refocusing capabilities of the model, we introduce MMVP \citep{tong2024eyes} and ${V}^{*}$ \citep{wu2024v} for fine-grained perception and spatial reasoning, where ${V}^{*}$ specifically targets visual grounding within complex environments. For complex logic and robustness, we include MMStar \citep{chen2024we} to assess high-quality visual reasoning and MMMU-Pro \citep{yue2025mmmu} to examine multi-step logical inference in complex scenarios with significant modality gaps. Finally, we incorporate POPE \citep{li2023evaluating} to measure model hallucination, ensuring a rigorous and thorough evaluation. Evaluation metrics include answer accuracy (Acc.) and the number of tokens generated when producing a correct answer (\# Tokens).

\subsection{Baselines and Implementation Details}
We compare LaRe with seven representative methods, including standard Chain of Thought reasoning methods that generate final answers through explicit step by step reasoning, text only multimodal reasoning approaches such as MM-CoT \citep{zhang2023multimodal} and CCoT \citep{mitra2024compositional}, Thinking with Images paradigms including ICoT \citep{gao2025interleaved} and CoF \citep{zhang2025chain}, and latent space reasoning paradigms including MCOUT \citep{pham2025multimodal} and LVR \citep{li2025latent}. 

Both the proposed method and the baselines are built upon the Qwen3-VL-2B/4B-Instruct \citep{bai2025qwen3} backbones. To ensure a fair comparison, we conduct single-stage supervised fine-tuning on these models using the Visual-CoT \citep{shao2024visual} dataset, and only the post-training stages for CoF and LVR follow their original configurations. Throughout the training procedure, the visual encoder and the multimodal projector remain frozen, and only the parameters of the LLM are updated. More implementation details for LaRe are provided in the Appendix~\ref{sec:details}.

\begin{table*}[h]
\centering
\renewcommand{\arraystretch}{1.0} 
\setlength{\tabcolsep}{6.5pt} 

\caption{Comparison to state-of-the-art baselines. The best results are \textbf{bolded} and the second best results are \underline{underlined}.}
\label{tab:larger}

\begin{adjustbox}{width=0.9\textwidth}
\begin{tabular}{lccccccc}
\toprule
Method & MMBench & MMStar & MMVP & POPE & ScienceQA & ${V}^{*}$ & MMMU-Pro \\
\midrule

Gemini2.5-Pro \citep{comanici2025gemini} & 84.2 & 74.7 & 45.5 & -- & -- & 81.3 & 59.6 \\
o3 \citep{openai2025o3} & -- & \underline{75.3} & -- & -- & -- & 82.7 & -- \\

\midrule

Janus-pro-7B \citep{chen2025janus} & 79.2 & 41.0 & 30.7 & 87.4 & -- & 65.6 & 38.1 \\
InternVL3.5-8B \citep{wang2025internvl3} & 79.5 & 69.3 & 50.6 & \underline{88.7} & 79.3 & 66.2 & 50.7 \\
Qwen3-VL-8B-Thinking \citep{bai2025qwen3} & \textbf{87.5} & 75.2 & -- & -- & 85.7 & 77.5 & 60.4 \\

\midrule

\multicolumn{8}{c}{\textit{Base VLM: Qwen3-VL-8B-Instruct}}\\
\hline

Vision-R1 \citep{huang2025vision} & 82.1 & 72.6 & 47.8 & 86.4 & \textbf{91.5} & 71.4 & 57.6 \\
Thyme \citep{zhang2025thyme} & 87.2 & 65.9 & 66.7 & 88.5 & 88.9 & 85.2 & \underline{64.8} \\
CapImagine \citep{li2026imaginationhelpsvisualreasoning} & 86.8 & 74.0 & 72.4 & 86.3 & 90.2 & \underline{87.9} & 59.4 \\
LVR \citep{li2025latent} & 83.2 & 67.1 & \underline{72.7} & 85.8 & 88.3 & 87.5 & 60.1 \\
\textbf{LaRe (Ours)} & \underline{87.4} & \textbf{77.1} & \textbf{74.0} & \textbf{89.6} & \textbf{91.5} & \textbf{92.7} & \textbf{67.1} \\

\bottomrule
\end{tabular}
\end{adjustbox}
\end{table*}

\subsection{Main Results}
We evaluate LaRe on multiple multimodal benchmarks, comparing it against existing explicit reasoning methods and recent latent space reasoning approaches. Table \ref{tab:main} summarizes the results.

\paragraph{Improved Accuracy.}
LaRe achieves the highest accuracy on most benchmarks. At the 2B scale, it attains 82.5\% on ScienceQA and 61.2\% on MMStar, outperforming competitive baselines including CoF and LVR. While CoF maintains a slight advantage on MMBench within the 2B backbone setting, LaRe demonstrates superior robustness on perception-intensive tasks such as MMVP and ${V}^{*}$, surpassing the runner-up by approximately 3.2 and 3.9 percentage points respectively. Specifically, LaRe surpasses latent reasoning baselines (MCOUT, LVR) by introducing dynamic refocusing for complex perception. Unlike image-injection methods like CoF, LaRe operates within the latent space to precisely capture fine-grained evidence, mitigating information loss from the modality gap.

\paragraph{Token Efficiency.}
LaRe demonstrates significant advantages in token efficiency. Unlike standard CoT or MM-CoT methods that typically require more than 150 tokens for inference, LaRe consistently maintains a token count below 100. For instance, on the POPE benchmark with a 2B backbone, LaRe utilizes only 18.9 tokens, which represents a nearly fivefold reduction compared to the 104.2 tokens consumed by standard CoT. This trend persists at the 4B scale, where LaRe achieves higher token efficiency than other latent space methods such as MCOUT and LVR while preserving high accuracy. The reduction in token usage decreases operational costs while enhancing the efficiency of model inference. This advancement offers significant benefits for real-world applications that prioritize cost, speed, and scalability.

\begin{figure}[h]
    \centering
    \includegraphics[width=1.0\linewidth]{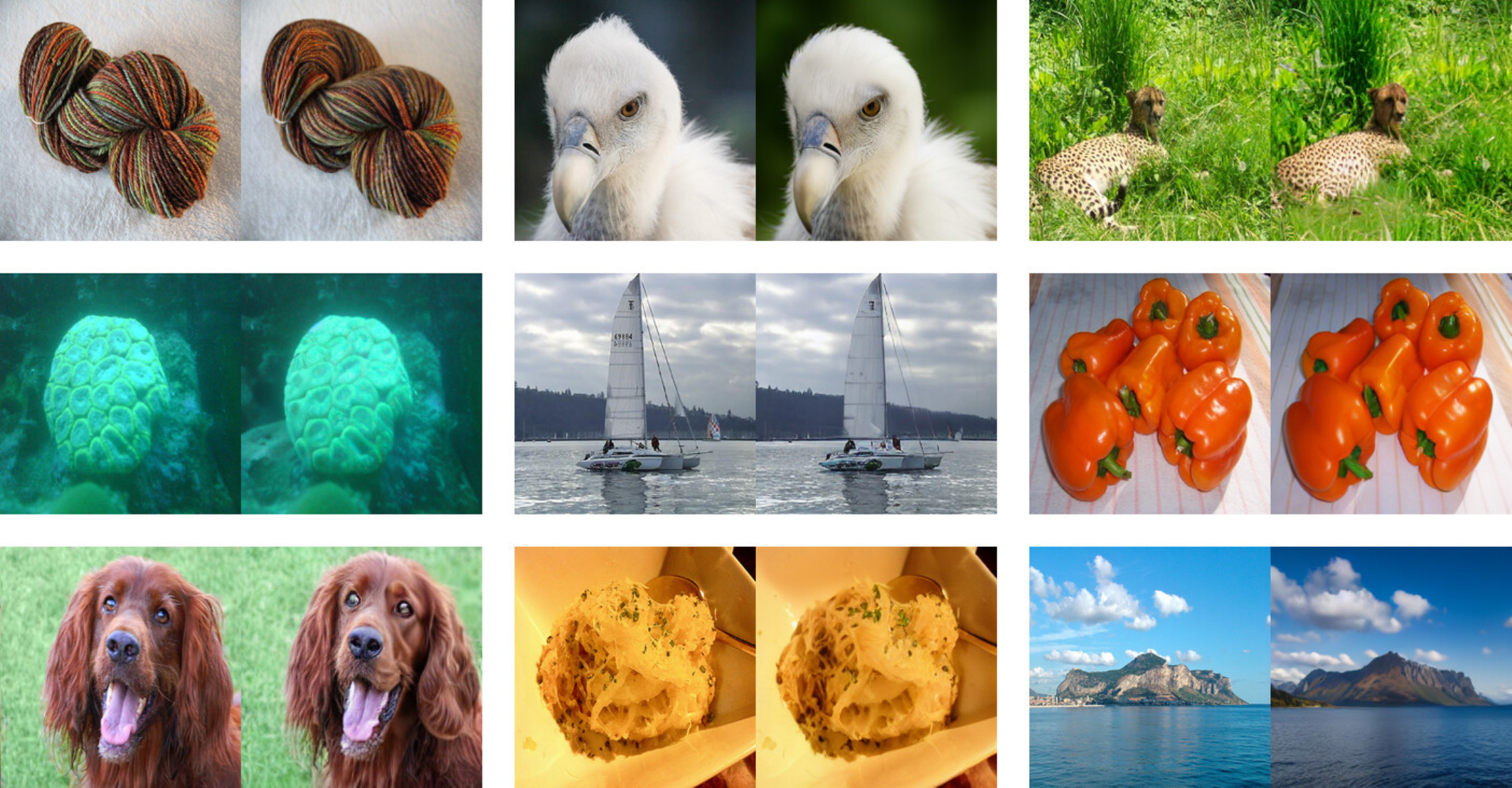}
    \caption{Visual reconstructions from latent sequences. In each panel, the left image shows the original input (from the ImageNet-1K validation set), and the right image shows its reconstruction.}
    \label{fig:recon}
\end{figure}

\subsection{Comparison With State-of-the-Arts}
Illustrated in Table~\ref{tab:larger}, we compare our LaRe based on Qwen3-VL-8B-Instruct~\citep{bai2025qwen3} with both private models~\citep{comanici2025gemini,openai2025o3}, open-sourced alternatives~\citep{chen2025janus,wang2025internvl3,bai2025qwen3}, and some multimodal reasoning methods~\citep{huang2025vision,li2026imaginationhelpsvisualreasoning,zhang2025thyme,li2025latent}. Experimental results demonstrate that LaRe achieves superior performance across various challenging multimodal benchmarks and surpasses existing state-of-the-art baselines. On the MMMU-Pro task which evaluates complex visual reasoning, LaRe not only outperforms open-source reasoning models of a similar scale such as Qwen3-VL-8B-Thinking but also exceeds the performance of larger closed-source models like Gemini 2.5 Pro and o3. Moreover, in comparison with other approaches based on the same VLM backbone such as Thyme and LVR, the performance lead of LaRe on the $V^*$ benchmark confirms that the latent refocusing mechanism effectively extracts fine-grained visual evidence and mitigates the modality gap.

\subsection{Qualitative Results}

\paragraph{Visualization of Each Iteration.}

\begin{figure*}[htbp]
    \centering
    \includegraphics[width=1.0\linewidth]{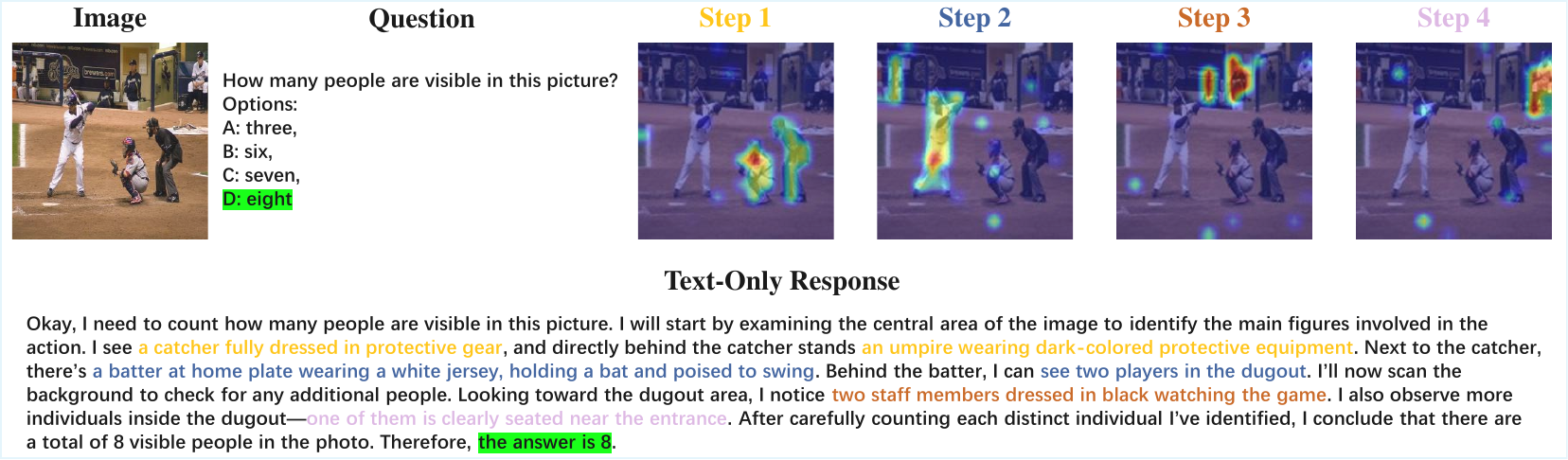}
    \caption{The evolution of attention in LaRe over four iterations and its alignment with the text-only reasoning trajectory.}
    \label{fig:analysis_step}
\end{figure*}

To intuitively illustrate the refocusing process of LaRe, Figure \ref{fig:analysis_step} presents a side by side visualization of the attention heatmaps produced by the Extractor module and the corresponding reasoning text trajectories across four consecutive iterations. We observe a highly consistent coevolution between the model’s visual focus regions and the process of textual evidence accumulation. In the first step, attention is primarily concentrated on the most salient foreground figures: the catcher and the umpire, which corresponds to the text marked in yellow. As reasoning progresses, the model refocuses on secondary regions: Step 2 localizes the batter and the players inside the dugout, corresponding to the blue text. In Steps 3–4, the model further refocuses on relatively concealed figures in the background, such as staff members dressed in black (corresponding to the brown text) and individuals seated near the entrance of the dugout (corresponding to the purple text). Through this sequential refocusing, the model arrives at the correct answer, indicated by the green text. This progressive transition from coarse grained salient targets to fine grained background details clearly indicates that LaRe does not perform a static single pass recognition, but instead implements a structured multi step evidence focusing process. Crucially, a high degree of synchronization is observed between the attention heatmaps and the color coded reasoning text. Each segment of textual reasoning, such as “two staff members dressed in black watching the game,” corresponds to the visual region emphasized at the current step.

\begin{table}[t]
\centering
\renewcommand{\arraystretch}{1.2} 
\setlength{\tabcolsep}{4.0pt} 
\scriptsize 
\caption{Ablation study of LaRe on Qwen3-VL-4B-Instruct backbone. The results demonstrate that each component contributes critically to the full model's performance and efficiency.}
\label{tab:ablation}
\begin{adjustbox}{width=\columnwidth}
\begin{tabular}{l|cc}
\bottomrule
Methods & Acc. & \# Tokens \\
\hline
LaRe & 74.7  & 65.0\\
 w/o Extractor & 69.3 (\textcolor{red}{-5.4}) & 89.1 (\textcolor{red}{+24.1}) \\
 w/o Semantic Augmentation & 67.2 (\textcolor{red}{-7.5}) & 68.5 (\textcolor{red}{+3.5}) \\
 w/o Extractor \& Semantic Augmentation & 64.5 (\textcolor{red}{-10.2}) & 95.9 (\textcolor{red}{+30.9}) \\

\toprule
\end{tabular}
\end{adjustbox}
\end{table}

\begin{table}[ht]
\centering
\renewcommand{\arraystretch}{1.0} 
\setlength{\tabcolsep}{7.0pt} 
\caption{Ablation results on Semantic Augmentation strategies. "Feature" refers to visual tokens from the image tokenizer, and "Pixel" refers to raw image pixels.}
\label{tab:semantic_ablation}

\fontsize{7pt}{8.4pt}\selectfont 

\begin{tabular}{lcccc} 
\toprule
Variant & MMStar & MMMU-Pro & MMVP & Avg. \\ \midrule 
\rowcolor[HTML]{EFEFEF} 
\textbf{LaRe} & \textbf{72.8} & \textbf{58.4} & \textbf{57.6} & \textbf{62.9}\\ 
Regression-Pixel & 69.2 & 55.1 & 52.4 & 58.9 \\
Regression-Feature & 70.8 & 56.4 & 54.1 & 60.4\\
Denoising-Pixel & 71.5 & 57.2 & 55.9 & 61.5\\ \midrule
w/o $\mathcal{L}_{rec}$ & 68.5 & 54.6  & 51.8 & 56.6\\ \bottomrule
\end{tabular}
\end{table}

\paragraph{From Latent to Images.}
Through image reconstruction visualization in Figure \ref{fig:recon}, we show that a U-net fine-tuned for five epochs on ImageNet-1K \citep{deng2009imagenet} can successfully recover clear scenes from the latent sequences. This result provides direct evidence that the Latent Refocusing process indeed focuses on visual evidence throughout the entire reasoning process.

\subsection{Ablation Study}

\paragraph{Role of Each Module.}
As shown in Table \ref{tab:ablation}, the full LaRe configuration achieves the most competitive performance, striking a superior balance between reasoning accuracy (74.7\%) and inference efficiency (65.0 tokens). The removal of Semantic Augmentation results in the sharpest decline in accuracy (-7.5\%), highlighting its indispensable role in regularizing the latent space and preserving the fine-grained visual-semantic fidelity required for complex logic. Meanwhile, discarding the Extractor leads to a 5.4\% drop in accuracy and a massive surge in token consumption (+24.1), demonstrating that the iterative refocusing mechanism is vital for pinpointing reasoning-relevant visual cues and preventing the model from generating verbose, redundant outputs. Finally, the simultaneous removal of both components leads to a compounded performance degradation, with accuracy falling to 64.5\% and token usage spiking to 95.9, which validates the synergy between latent-space structuring and dynamic refocusing in our proposed paradigm.

\paragraph{Impact of Reconstruction Objective.}
Table \ref{tab:semantic_ablation} presents the impact of different reconstruction objectives within the semantic augmentation strategy. Experimental results indicate that the removal of the reconstruction loss $\mathcal{L}_{rec}$ leads to a significant performance decline across all benchmarks, where the average accuracy drops from 62.9\% to 56.6\%. This observation underscores the role of the visual reconstruction task in guiding the model to establish structured latent space representations, which ensures that latent variables retain essential visual evidence throughout multiple reasoning iterations. Moreover, the feature-space denoising reconstruction mechanism utilized in LaRe demonstrates a clear advantage over ablation variants based on raw pixels or the standard regression objective. This suggests that the implicit regularization introduced by the denoising task allows the model to capture high-level spatial logic and fine-grained semantics more effectively, which substantially improves the reliability of multimodal reasoning.

\begin{figure}[htbp]
    \centering
    \includegraphics[width=1.0\linewidth]{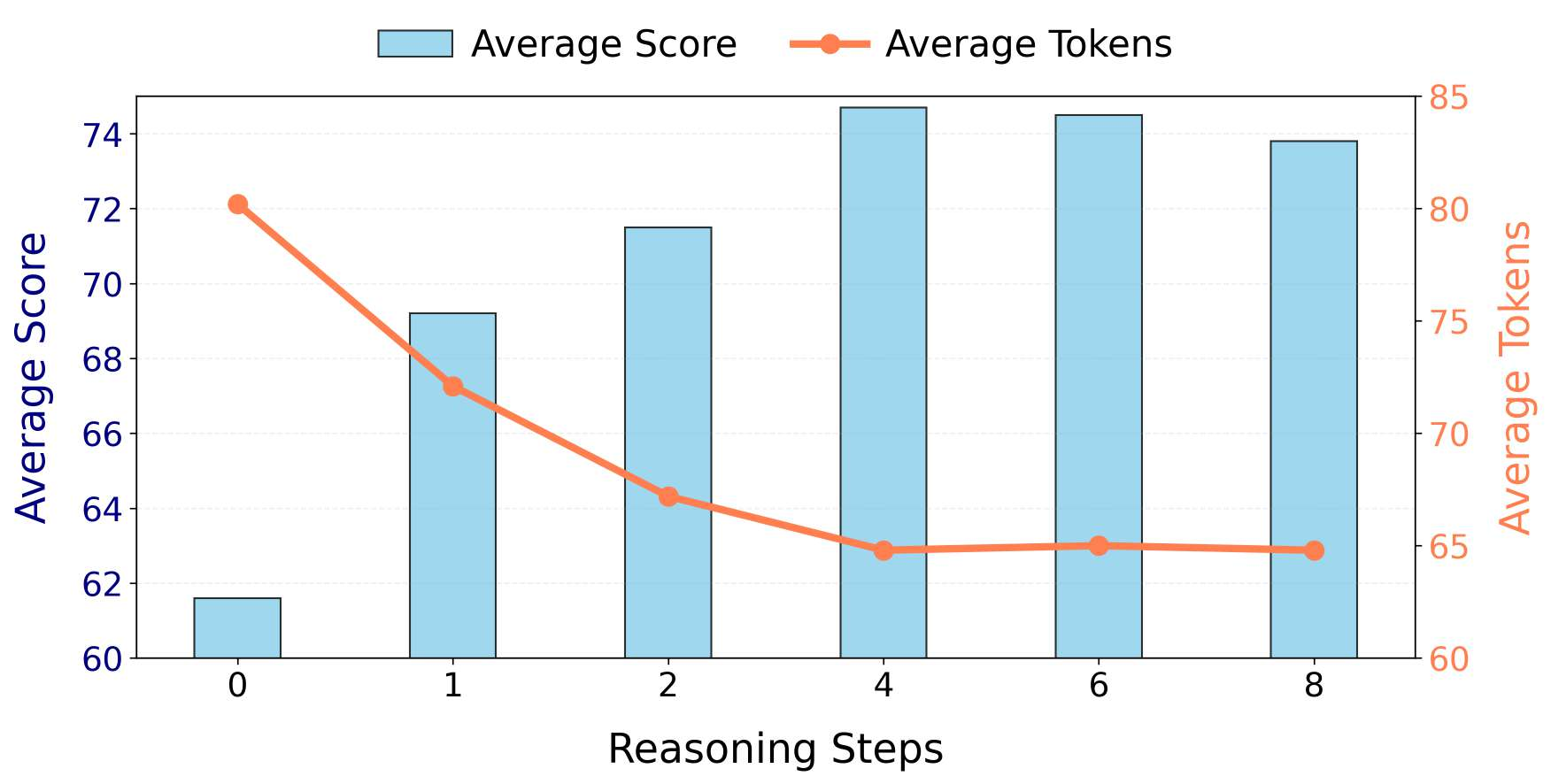}
    \caption{Impact of varying the number of reasoning steps on model performance and output efficiency.}
    \label{fig:step_abl}
\end{figure}

\begin{table}[H]
    \centering
    \renewcommand{\arraystretch}{1.0} 
    \setlength{\tabcolsep}{2.5pt} 
    \footnotesize 
    \caption{Quantitative comparison on attention values. The results indicated that LaRe achieves higher values than other latent reasoning methods.}
    \label{tab:attention_comparison}
    \begin{adjustbox}{width=\columnwidth}
    \begin{tabular}{l c c c c}
    \toprule
    Statistic ($\times 10^{-4}$) & MCOUT & LVR & LaRe & P-value \\
    \midrule
    Mean & 4.32 & 7.34 & 10.94 & $1.07 \times 10^{-7}$ \\
    25th Percentile & 3.90 & 6.63 & 10.06 & -- \\
    Median & 4.29 & 7.33 & 10.87 & $3.79 \times 10^{-9}$ \\
    75th Percentile & 4.71 & 8.01 & 11.79 & -- \\
    95th Percentile & 5.36 & 9.11 & 13.16 & -- \\
    \bottomrule
    \end{tabular}
    \end{adjustbox}
\end{table}

\paragraph{Impact of Reasoning Steps.}
The ablation study on the number of iterative steps (Figure \ref{fig:step_abl}) reveals a clear pattern of diminishing returns. Increasing the initial number of steps markedly improves performance and substantially reduces output length, validating the effectiveness of the iterative refocusing process. However, once the number of steps reaches a saturation threshold, both performance gains and output compression plateau. This finding indicates that the model can attain a stable reasoning state and offers guidance on the optimal number of steps to balance computational cost and reasoning effectiveness.

\paragraph{Attention Analysis.}
We compute the mean attention scores from all latent tokens to all visual tokens using the MMStar dataset. We employ T-test\citep{student1908probable} to compare the means and the Mann-Whitney U test\citep{mann1947test} to compare the medians of the two distributions. As shown in Table~\ref{tab:attention_comparison}, LaRe achieves higher attention scores than other latent reasoning methods. This indicates that LaRe effectively guides the model to focus more on the input visual signals. Extended experiments on the differences between LaRe and other latent reasoning approaches can be found in Table~\ref{tab:vaed} of Appendix~\ref{subsec:anal}.

\section{Conclusion}
In this work, we introduced LaRe, a multimodal reasoning paradigm that bridges the modality gap by performing visual refocusing within a latent space. By decoupling the reasoning process from explicit token generation, LaRe addresses the limitations of "Thinking with Images" while maintaining dynamic flexibility through its Latent Refocusing Mechanism. Our semantic augmentation training strategy further ensures that the latent space is semantically structured and grounded in fine-grained visual evidence. Experimental evaluations demonstrate that LaRe outperforms current baselines in accuracy and reduces inference token consumption by an average of 59.7\%. Scaling to larger backbones further confirms its effectiveness and robustness. These results highlight the potential of latent refocusing as a more efficient alternative for complex multimodal tasks.

\section*{Limitations}
A primary limitation of the current LaRe framework is its reliance on a predefined, static number of reasoning iterations $K$. While our empirical analysis identifies an optimal saturation threshold for performance, this "one-size-fits-all" approach does not account for the varying complexity across different multimodal tasks. For relatively straightforward visual queries, a fixed number of latent iterations may introduce redundant computational overhead and unnecessary latency. Conversely, for exceptionally intricate logical reasoning tasks, the model's performance may still be constrained by the preset iteration limit. Developing a dynamic halting mechanism, which allows the model to adaptively determine the required number of latent thoughts based on the evolving reasoning state, remains an important direction for further enhancing the efficiency and flexibility of the LaRe paradigm.

\section*{Ethical Considerations}

The datasets utilized in this study are publicly available benchmarks widely recognized in the research community. These datasets consist of images and textual annotations sourced from public domains and are used strictly for academic evaluation. To the best of our knowledge, all visual and textual data have been curated to ensure they do not contain Personally Identifiable Information (PII) or sensitive personal content. Our experimental procedures strictly adhere to the licensing agreements of the respective datasets and align with the ethical guidelines for responsible artificial intelligence research. Therefore, we believe that this work fully complies with the ethical standards of the conference.

\FloatBarrier

\bibliography{custom}

\clearpage
\appendix
\section*{Appendix}

\begin{algorithm}[H]
\caption{Latent Refocusing (LaRe)}
\label{alg:LaRe}
\begin{algorithmic}[1]
\Require Input image $i$, text prompt $t$, number of reasoning steps $K$, special tokens $\texttt{<|bot|>}$ \& $\texttt{<|eot|>}$
\Ensure Final answer ${Y}$

\State ${V}_T \gets \text{Projector}(\text{VisualEncoder}(i))$ \Comment{Extract visual tokens}
\State ${T} \gets \text{Embed}(t)$ \Comment{Project text to embedding space}

\State ${Z} \gets \emptyset$ \Comment{Initialize latent token sequence}

\For{$k = 1$ \textbf{to} $K$}
    \State ${H}_T^{(k)} \gets {H}_{\text{LLM}}^{(k)}([{V}_T; {T}; {Z}])$ \Comment{Extract current reasoning state from LLM}
    \State ${Z}^{(k)} \gets \text{Extractor}({H}_T^{(k)}, {V}_T)$ \Comment{Refocus on ${V}_T$ and generate new latent token}
    \State ${Z} \gets [{Z}; {Z}^{(k)}]$ \Comment{Update the latent thought chain}
\EndFor

\State $\text{Input}_{final} \gets [{V}_T; {T}; \texttt{<|bot|>}; {Z};\texttt{<|eot|>}]$ \Comment{Concatenate all modalities}
\State ${Y} \gets \text{LLM}_{\text{generate}}(\text{Input}_{final})$ \Comment{Autoregressive answer generation}

\State \Return ${Y}$
\end{algorithmic}
\end{algorithm}

\section{Implementation Details}\label{sec:details}

\paragraph{Architecture of Extractor.}

\begin{figure}[htbp]
\centering
\includegraphics[width=0.5\linewidth]{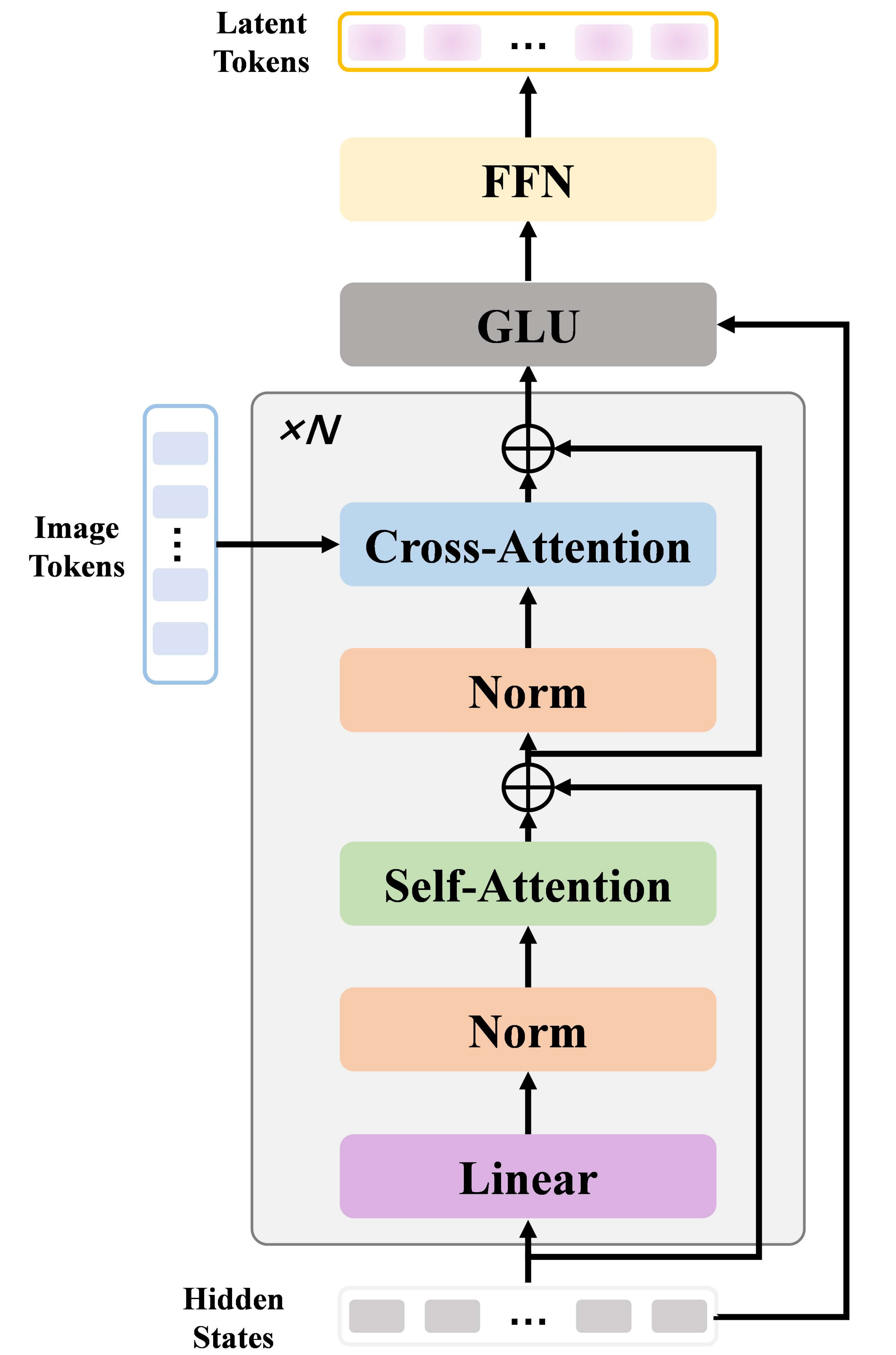}
\caption{The detailed architecture of the Extractor. The module iteratively refines features through self-attention and cross-attention layers before passing them through a GLU and FFN.}
\label{fig:extractor}
\end{figure}

The Extractor module serves as the core component for executing iterative refocusing and adopts a Transformer-like architectural design, as shown in figure~\ref{fig:extractor}. It consists of 2 iterative layers. In each layer, features are first preprocessed through a linear transformation, followed by a multi-head self-attention mechanism with 12 attention heads to model the evolution of internal reasoning dynamics. On this basis, a cross-attention mechanism, also configured with 12 heads, is applied over visual features to perform precise refocusing, thereby extracting evidence that is highly relevant to the current reasoning step. To ensure sufficient representational capacity, the hidden dimension of the Extractor is uniformly set to 1536. After completing the iterative process, the refined features are sequentially passed through a gated linear unit (GLU) and a feed-forward network (FFN), ultimately producing a latent representation that drives subsequent reasoning.

\paragraph{Settings of Image Reconstruction.}
The diffusion process employs a linear noise schedule defined by $\beta_t \in [\beta_{start}=1\times10^{-4}, \beta_{end}=0.02]$ over $T = 1000$ iterations. At each training step, a random timestep $t \sim U(1, T)$ is sampled, and the input image $x_0$ is corrupted as $x_t = \sqrt{\bar{\alpha}_t} x_0 + \sqrt{1 - \bar{\alpha}_t} \epsilon$, where $\bar{\alpha}_t = \prod_{s=1}^t (1 - \beta_s)$. The sampling process follows an iterative denoising procedure using the predicted noise and the reparameterization of the posterior distribution. We adopt KL-16 \citep{rombach2022high} as the image tokenizer, which is a continuous VAE \citep{kingma2013auto} with a Kullback–Leibler (KL) divergence regularization term. The architecture of our denoising U-Net is illustrated in Table~\ref{tab:unet_architecture}.

\begin{table*}[htbp]
\centering
\small
\setlength{\tabcolsep}{8pt}
\caption{U-Net configuration.}
\label{tab:unet_architecture}
\begin{tabular}{@{} l l @{}}
\toprule
Parameter & Value \\
\midrule
sample\_size & \texttt{img\_size} (input spatial size) \\
in\_channels & \texttt{self.channel} (input feature channels) \\
out\_channels & \texttt{self.channel} (output feature channels) \\
layers\_per\_block & 3 \\
block\_out\_channels & (96, 192, 384, 512) \\
down\_block\_types & (\texttt{DownBlock2D}, \texttt{CrossAttnDownBlock2D}, \texttt{CrossAttnDownBlock2D}, \texttt{DownBlock2D}) \\
up\_block\_types & (\texttt{UpBlock2D}, \texttt{CrossAttnUpBlock2D}, \texttt{CrossAttnUpBlock2D}, \texttt{UpBlock2D}) \\
cross\_attention\_dim & 768 \\
attention\_head\_dim & 64 \\
norm\_num\_groups & 32 \\
\bottomrule
\end{tabular}
\end{table*}

\paragraph{Other Hyperparameters.}
We set the number of iterations $K$ to 4 for LaRe. The training process employs the AdamW optimizer \citep{loshchilov2017decoupled} with $\beta_1 = 0.9$, $\beta_2 = 0.95$ and weight decay of $0.05$. The loss coefficients are set to $\lambda_{rec} = 1.0$ for the semantic augmentation, determined through grid search on the validation set. All models are trained using mixed precision (FP16) on eight NVIDIA A100 80GB GPUs with a global batch size of 128. To address GPU memory limitations, we implement gradient accumulation over four steps, which results in a micro-batch size of 4 per GPU. The memory workload is uniformly distributed across all devices through the integration of DeepSpeed ZeRO-3.

\paragraph{Image Resolution Configurations}
For image resolution configurations, all input images are resized to $224\times224$ pixels to match the standard input requirements of the CLIP-ViT-L/14 visual encoder. This resolution yields 256 visual tokens after processing through the vision encoder (a $16\times16$ grid of patch tokens). The diffusion-based reconstructor operates on latent representations with a feature map shape of $[4,28,28]$, corresponding to the compressed spatial dimensions while preserving essential visual information. During reconstruction evaluation, the output images are upsampled back to the original $224\times224$ resolution for qualitative assessment.

\paragraph{Settings of Probing Analysis.} 
To rigorously evaluate whether the latent tokens ${Z}$ effectively encode joint linguistic reasoning and visual semantics, we conducted a comprehensive probing analysis following established methodologies in representation learning. The pooled latent vectors ${z}_{\text{agg}} = \phi_{\text{pool}}({Z})$, textual rationale embeddings ${r} = \phi_{\text{pool}}({R})$, and visual anchors ${v} = \phi_{\text{pool}}({V})$ were computed using the same hybrid pooling module (mean-max pooling) employed during training. All representations were subsequently projected to a 256-dimensional space via a lightweight, trainable linear layer and L2-normalized prior to classification to ensure scale-invariant comparison. For binary classification, positive pairs were constructed from matching (image, question) instances, while negative pairs were sampled in a balanced manner from three distinct strategies to prevent trivial solutions: (i) different-question/same-image, (ii) different-image/same-question, and (iii) random in-batch negatives. The final probing dataset comprised 1,000 labeled pairs (500 positive/500 negative) with a stratified 70/30 train/test split based on negative sampling type. We employed two distinct probe architectures: a Linear Support Vector Machine (SVM) with a linear kernel optimized via 5-fold cross-validation on the training set, and a single-layer Multi-Layer Perceptron (MLP) implemented in PyTorch consisting of one hidden layer, optimized with Adam ($\text{lr}=1\times10^{-3}$) and early stopping based on validation loss.

\section{Some Supplement Experiments Results}\label{sec:more_exp}

\subsection{More Ablations}

\begin{table}[ht]
\centering
\caption{Ablation study on Extractor design. "Dynamic" indicates using $H_{T}^{(k)}$ as the query source. All variants are evaluated on the 4B backbone.}
\label{tab:extractor_ablation}
\small 
\begin{adjustbox}{width=\columnwidth}
\begin{tabular}{@{}llcc@{}}
\toprule
Variant & Component & Acc. & \# Tokens \\ \midrule
\rowcolor[HTML]{F2F2F2} 
\textbf{LaRe} & \textbf{Full} & \textbf{74.7} & \textbf{65.0} \\ 
w/o Self-Attn & Only Cross-Attn & 71.1 & 72.7 \\
Linear Proj. & Linear Bottleneck & 66.5 & 82.3 \\
GAP & Global Avg Pooling & 67.8 & 80.1 \\ \midrule
Fixed Query & Learnable Latents & 72.9 & 67.5 \\ \bottomrule
\end{tabular}
\end{adjustbox}
\end{table}

\paragraph{Ablation Study on Extractor Architecture.}

We investigate the architectural design of the Extractor through a series of ablation experiments on a 4B parameter backbone, as summarized in Table \ref{tab:extractor_ablation}. The results demonstrate that the full LaRe configuration provides an optimal trade-off between inference accuracy (74.7\%) and token consumption (65.0 tokens). Specifically, the omission of the self-attention mechanism within the Extractor leads to a significant 3.6\% decrease in accuracy alongside an increase in the number of inference tokens, which confirms that logic modeling within the latent space is critical for maintaining reasoning coherence. Furthermore, the performance degradation observed when replacing the Transformer structure with simpler alternatives, such as a linear projection layer or global average pooling, indicates that basic bottleneck structures are insufficient for capturing the complex high-dimensional mappings between multimodal features, as accuracy drops to 66.5\% and 67.8\% respectively. Finally, a comparison with a fixed query baseline shows that employing learnable latent variables instead of dynamic queries driven by $H_{T}^{(k)}$ results in a 1.8\% reduction in performance. This finding supports the central design principle that dynamic refocusing based on the current reasoning state is essential for the precise extraction of visual evidence.

\begin{table}[t]
\centering
\renewcommand{\arraystretch}{1.2} 
\setlength{\tabcolsep}{2.2pt} 
\footnotesize 
\caption{Performance comparison of adaptive reasoning mechanisms using different convergence criteria.}
\label{tab:adaptive}
\begin{adjustbox}{width=\columnwidth}
\begin{tabular}{l|ccc}
\bottomrule
Methods & Average Steps & Acc. & \# Tokens \\
\hline
LaRe & 4.0 & 74.7  & 65.0\\
+ Adaptive (Cosine) & 5.9 & 74.5 & 67.8 \\
+ Adaptive (L1 Distance) & 2.8 & 73.1 & 71.3 \\
+ Adaptive (L2 Distance) & 2.9 & 74.2 & 70.5 \\
\toprule
\end{tabular}
\end{adjustbox}
\end{table}

\begin{table*}[t]
\centering
\caption{Ablation study of different Image Tokenizers for Semantic Augmentation. All variants are trained using the same denoising reconstruction objective. The results demonstrate that KL-16 provides the most effective latent representations for multimodal reasoning.}
\label{tab:tokenizer_ablation}
\begin{adjustbox}{width=0.6\textwidth}
\begin{tabular}{lccccc}
\toprule
Tokenizer & Pre-train Task & MMStar & MMMU-Pro & MMVP & Avg. \\ \midrule
MAE & Generative & 69.2 & 53.5 & 55.1 & 59.3 \\
CLIP-L/336 & Contrastive & 72.5 & 55.8 & 51.2 & 59.2 \\
DINOv2 & Discriminative & 71.1 & 54.2 & 55.8 & 59.7 \\
SigLIP-L/384 & Contrastive & \textbf{72.8} & 56.9 & 54.6 & 61.0 \\ \midrule
\rowcolor[HTML]{EFEFEF} 
\textbf{KL-16 (Ours)} & \textbf{Hybrid/Conv} & \textbf{72.8} & \textbf{58.4} & \textbf{57.6} & \textbf{62.9} \\ \bottomrule
\end{tabular}
\end{adjustbox}
\end{table*}

\paragraph{Choices of the Image Tokenizer.}

To evaluate the influence of image tokenizer selection within the semantic augmentation strategy on reasoning performance, Table \ref{tab:tokenizer_ablation} presents a comparison of model performance across various pre-training paradigms. Experimental results indicate that KL-16, which utilizes a hybrid convolutional architecture, outperforms other variants across all metrics. It achieves an average accuracy of 62.9\%, which markedly exceeds the generative task based MAE (59.3\%), the discriminative DINOv2 (59.7\%), and the contrastive learning model SigLIP (61.0\%). In particular, KL-16 demonstrates a substantial lead on the MMMU-Pro (58.4\%) and MMVP (57.6\%) benchmarks, both of which require sophisticated visual logic. These results verify that the latent representations from KL-16 maintain a superior balance between high-level semantic consistency and low-level visual details during the denoising reconstruction process compared to pure semantic or pixel-level objectives. As a result, this provides high-quality supervision signals for the latent refocusing mechanism and enables the model to capture fine-grained visual evidence essential for multimodal reasoning.

\paragraph{Adaptive Reasoning Steps.}
To enhance efficiency, we investigate an adaptive stopping mechanism that determines the termination of reasoning based on the convergence of latent tokens. We evaluate this mechanism by tracking token evolution throughout the iterative process using cosine similarity, $L_{1}$ distance, and $L_{2}$ distance. As illustrated in Table~\ref{tab:adaptive}, these metrics reveal a convergence trend where inter-step distances decrease monotonically as reasoning progresses. This observation confirms the intrinsic stability of latent reasoning and supports the conclusions drawn from the fixed-step ablation study in Figure~\ref{fig:step_abl}. However, the results also demonstrate a significant trade-off between efficiency and performance. While adaptive stopping potentially mitigates redundancy and excessive computation, it results in a slight decline in overall performance metrics. We hypothesize that these heuristic adaptive stopping mechanisms are insufficient for accurately estimating the number of steps required for reasoning, and may instead lead to underthinking or overthinking. Consequently, future research focuses on developing more refined stopping criteria to achieve truly efficient adaptive inference without compromising model performance.

\begin{table}[t]
\centering
\scriptsize 
\caption{Performance comparison of different $\beta$ schedules in diffusion-based reconstruction.}
\label{tab:beta_schedule}
\begin{adjustbox}{width=\columnwidth}
\begin{tabular}{l|cc}
\bottomrule
Methods & Avg. Score & Avg. Tokens \\
\hline
Linear ($\beta_{\text{max}}=0.02$) & 74.7 & 65.0 \\
Linear ($\beta_{\text{max}}=0.05$) & 73.9 & 65.7 \\
Cosine ($\beta_{\text{max}}=0.02$) & 74.7 & 65.2 \\
Cosine ($\beta_{\text{max}}=0.05$) & 74.1 & 64.8 \\
Squared Cosine ($\beta_{\text{max}}=0.02$) & 72.4 & 66.1 \\
Squared Cosine ($\beta_{\text{max}}=0.05$) & 73.6 & 65.3 \\
\toprule
\end{tabular}
\end{adjustbox}
\end{table}

\paragraph{Schedule of $\beta$ in Image Reconstruction.}
We investigate the sensitivity of Semantic Augmentation to the noise schedule $\beta$ used in the diffusion-based reconstruction objective. As shown in Table~\ref{tab:beta_schedule}, we compare linear, cosine, and squared cosine schedules across different maximum $\beta$ values. The cosine schedule with $\beta_{\text{max}}=0.02$ achieves the best balance between reconstruction quality and reasoning performance, yielding the highest average score of 74.7\%. While linear schedules demonstrate competitive performance, they exhibit slightly higher variance across tasks. The squared cosine schedule shows comparable results but requires more careful tuning of the maximum $\beta$ value. These results indicate that LaRe is reasonably robust to different schedule choices, though smoother transitions in the noise schedule (as in cosine) appear to provide more stable training dynamics for the latent reasoning process.

\paragraph{On the choice of $l$.}
Setting $l$ to the number of LLM layers is not a purely heuristic choice. Previous work~\citep{skean2025layer} shows that different layers of an LLM encode semantically distinct representations. Inspired by this, our layer-aligned design assigns a dedicated latent token as a semantic anchor to the hidden state of each layer, enabling the extractor to capture reasoning cues at every level of abstraction. To further validate this design, we conduct an ablation study on $l$ using Qwen3-VL-2B-Instruct, which has 28 LLM layers, where hidden states are selected sequentially starting from the last layer. As shown in Table~\ref{tab:ablation_l}, $l = L$ achieves the highest accuracy with only negligible additional latency compared to smaller $l$, indicating that aligning the number of latent tokens with the number of LLM layers yields the best accuracy-efficiency trade-off.

\begin{table}[h]
\centering
\caption{Ablation study of the number of latent tokens $l$ on Qwen3-VL-2B-Instruct.}
\label{tab:ablation_l}
\begin{adjustbox}{width=\columnwidth}
\begin{tabular}{lccc}
    \toprule
    $l$ & Average Accuracy (\%) & Token Count & Latency (s/step) \\
    \midrule
    $L/4~(7)$  & 63.2          & 232          & \textbf{0.42} \\
    $L/2~(14)$ & 64.8          & 225          & 0.44 \\
    $L~(28)$   & \textbf{66.2} & \textbf{218} & 0.47 \\
    \bottomrule
\end{tabular}
\end{adjustbox}
\end{table}

\subsection{More Analysis}
\label{subsec:anal}

\paragraph{Quantitative Analysis of Visual Refocusing.}

Unlike existing latent-space reasoning methods such as LVR, our Extractor actively refocuses to the visual evidence at every latent step, rather than merely propagating a statically encoded visual context along the reasoning trajectory. Structurally, this distinction is already built into our design through the per-step cross-attention to $V_T$; the claim itself, however, has so far remained largely qualitative. To render it falsifiable and quantitatively measurable, this section introduces a concise yet diagnostic metric, the Visual Attention Entropy Drift (VAED), together with two complementary diagnostics, in order to directly verify whether the latent-space iterations of LaRe correspond to genuine visual refocusing.

Let $\alpha^{(k)} \in \mathbb{R}^{|V_T|}$ denote the normalized cross-attention distribution from the latent query to the visual token set at the $k$-th step, averaged over all attention heads and latent positions. We define VAED as the mean absolute change of attention entropy between consecutive reasoning steps:
\begin{equation}
\label{eq:vaed}
\text{VAED} \;=\; \frac{1}{K-1}\sum_{k=2}^{K} \Big| \, H\big(\alpha^{(k)}\big) - H\big(\alpha^{(k-1)}\big) \, \Big|,
\end{equation}
where $H(\cdot)$ denotes the Shannon entropy. The intuition behind this metric is straightforward: when VAED is close to zero, the visual attention distribution remains nearly invariant across reasoning steps, indicating that the latent state is merely forwarded without truly re-querying the visual evidence, which is precisely the failure mode we attribute to methods lacking a look-back mechanism; conversely, a VAED that is significantly greater than zero indicates that attention over visual tokens is meaningfully reallocated between consecutive steps, which is the behavioral signature of iterative visual refocusing. 

For baselines that do not explicitly expose per-step cross-attention, such as LVR, we extract an implicit attention distribution by computing the cosine similarity between each latent state and the visual token embeddings followed by a softmax normalization, thereby interpreting their attention behavior in the most permissive way to ensure a fair comparison. To rule out the alternative hypothesis that a high VAED arises merely from random attention reshuffling rather than purposeful refocusing, we additionally report two complementary diagnostics: the Top-5 coverage shift ($\Delta_{\text{top-5}}$), defined as the mean Jaccard distance between the Top-5 attended visual token sets across consecutive steps, which characterizes the spatial concreteness of the refocusing; and the average attention entropy $\bar{H}$ across all steps, which characterizes the selectivity of attention at each individual step.

\begin{table*}[t]
\centering
\small
\setlength{\tabcolsep}{6pt}
\renewcommand{\arraystretch}{1.15}
\begin{tabular}{lccccc}
\toprule
Method & Structural Refocusing? & VAED $\uparrow$ & $\Delta_{\text{top-5}}$ $\uparrow$ & $\bar{H}$ $\downarrow$ & Acc. (MMStar) \\
\midrule
Standard CoT & N/A (no latent step) & -- & -- & -- & 56.6 \\
MCOUT & $\times$ & 0.05 & 0.08 & 3.07 & 58.2 \\
LVR & $\times$ & 0.03 & 0.05 & 3.18 & 56.6 \\
MCOUT + Cross-Attn$^{\dagger}$ & Partial & 0.12 & 0.19 & 2.61 & 59.6 \\
\midrule
LaRe (Ours) & $\checkmark$ & \textbf{0.31} & \textbf{0.42} & \textbf{2.14} & \textbf{61.2} \\
\bottomrule
\end{tabular}
\caption{Quantitative analysis of visual refocusing behavior. Higher VAED and Top-5 shift indicate stronger look-back behavior, while lower $\bar{H}$ indicates more selective attention. $^{\dagger}$ Ablation variant that augments MCOUT with per-step cross-attention to $V_T$ but does not employ our diffusion-style reconstruction objective.}
\label{tab:vaed}
\end{table*}

Based on these metrics, we evaluate all methods on 200 instances randomly sampled from MMStar and V$^{*}$, with the Qwen3-VL-2B-instruct backbone held fixed, and summarize the results in Table~\ref{tab:vaed}. We observe a consistent and striking pattern: LaRe attains a VAED of 0.31, roughly 6$\times$ that of MCOUT (0.05) and 10$\times$ that of LVR (0.03), confirming that the attention distribution of LaRe indeed undergoes substantial migration across latent reasoning steps, whereas the attention of the baselines stays nearly stationary throughout the process. This drift is not a diffuse spread: the Top-5 coverage shift of LaRe reaches 0.42, whereas that of MCOUT is only 0.08, indicating that LaRe genuinely switches the image regions it attends to across reasoning steps, a spatially localizable signature of refocusing behavior rather than a numerical artifact. Crucially, this stronger refocusing behavior does not come at the cost of attention quality: despite the markedly larger drift, LaRe maintains a lower average attention entropy $\bar{H}$ (2.14, compared with 3.07 for MCOUT), showing that the attention at each step still concentrates sharply on a small number of relevant regions rather than diffusing into uninformative noise. This combination of higher drift coupled with lower per-step entropy is particularly important, since it directly excludes the most competitive alternative explanation, namely that a high VAED reflects nothing more than noisy random attention reshuffling, thereby supporting the interpretation that the latent-space iterations of LaRe correspond to purposeful, evidence-anchored migration of visual refocusing.

Taken together, these diagnostics provide direct quantitative evidence for the following conclusion: the performance gains of LaRe over existing latent-space reasoning methods originate from a structurally distinct mechanism, that of iterative and selective visual refocusing, rather than from incidental factors such as increased parameter count, the mere introduction of cross-attention, or regularization side-effects during training.

\begin{table*}[t]
\centering
\caption{Comparison of training computational overhead across different model scales.}
\label{tab:overhead}
\renewcommand{\arraystretch}{1.2} 
\setlength{\tabcolsep}{8pt}      
\begin{adjustbox}{width=0.7\textwidth}
\begin{tabular}{lcccc}
\toprule
Methods & Trainable Params & Speed (s/iter) & Time & GPU Memory \\
\midrule
\multicolumn{5}{c}{\textit{Base VLM: Qwen3-VL-2B-Instruct}} \\
\hline 
Standard CoT & 1.72 B & 2.62 & 12h 17min & 24.25 GB \\
+ Extractor & 1.73 B & 2.79 & 13h 12min & 25.57 GB \\
+ Semantic Aug. & 1.76 B & 2.71 & 12h 40min & 25.13 GB \\
\cellcolor[HTML]{F2F2F2}LaRe (Full) & \cellcolor[HTML]{F2F2F2}1.77 B & \cellcolor[HTML]{F2F2F2}2.88 & \cellcolor[HTML]{F2F2F2}13h 30min & \cellcolor[HTML]{F2F2F2}26.08 GB \\
\midrule
\multicolumn{5}{c}{\textit{Base VLM: Qwen3-VL-4B-Instruct}} \\
\hline
Standard CoT & 4.02 B & 4.59 & 21h 44min & 38.15 GB \\
+ Extractor & 4.03 B & 4.84 & 22h 56min & 40.21 GB \\
+ Semantic Aug. & 4.07 B & 4.73 & 22h 25min & 39.70 GB \\
\cellcolor[HTML]{F2F2F2}LaRe (Full) & \cellcolor[HTML]{F2F2F2}4.08 B & \cellcolor[HTML]{F2F2F2}4.96 & \cellcolor[HTML]{F2F2F2}23h 29min & \cellcolor[HTML]{F2F2F2}41.20 GB \\
\bottomrule
\end{tabular}
\end{adjustbox}
\end{table*}

\paragraph{Analysis of Computational Costs.}

To assess the computational feasibility of the LaRe paradigm, Table \ref{tab:overhead} provides a detailed comparison of training overhead across various model scales. The experimental results demonstrate that the computational burden introduced by the Extractor module and the reconstruction-based semantic augmentation remains marginal in comparison to the significant performance gains. In the case of the 2B backbone, the trainable parameters of the full LaRe model increase by approximately 2.9\%, expanding from 1.72B to 1.77B, while the total training duration extends by only 1.2 hours with an iteration speed maintained at 2.88 seconds per step. Regarding GPU memory utilization, LaRe requires an additional 1.83 GB for the 2B scale and 3.05 GB for the 4B scale compared to Standard CoT, which suggests high engineering adaptability for mainstream computing devices. Crucially, as the semantic augmentation module operates exclusively during the training phase and the latent-space iterations alleviate the computational demands of explicit text generation, this marginal training investment achieves a competitive balance between efficiency and effectiveness by enhancing inference accuracy and reducing token overhead.

\begin{table*}[htbp]
    \centering
    \renewcommand{\arraystretch}{1.1}
    \setlength{\tabcolsep}{5.5pt}
    \tiny
    \caption{Average inference latency (s/query) across different benchmarks. The total latency includes both latent reasoning and final text decoding. Lower is better.}
    \label{tab:latency}
    \begin{adjustbox}{width=0.7\textwidth}
    \begin{tabular}{@{} l | ccccccc | c @{}}
    \toprule
    Method & MMBench & MMStar & MMVP & ScienceQA & POPE & $V^*$ & MMMU-Pro & \textbf{Avg.} \\ 
    \midrule
    \multicolumn{9}{c}{\textit{Base VLM: Qwen3-VL-2B-Instruct}} \\
    \hline
    Standard CoT & 1.15 & 1.28 & 0.82 & 1.25 & 0.65 & 1.32 & 1.55 & 1.15 \\
    MM-CoT       & 1.12 & 1.15 & 0.78 & 1.32 & 0.62 & 1.28 & 1.45 & 1.10 \\
    CCoT         & 0.92 & 1.22 & 0.95 & 1.45 & 0.72 & 1.55 & 1.48 & 1.18 \\
    ICoT         & 1.05 & 1.18 & 0.85 & 1.22 & 0.68 & 1.42 & 1.62 & 1.15 \\
    CoF          & 1.18 & 1.35 & 0.92 & 1.38 & 0.85 & 1.82 & 1.95 & 1.35 \\
    MCOUT        & 0.58 & 0.72 & 0.48 & 0.65 & 0.32 & 0.75 & 0.92 & 0.63 \\
    LVR          & 0.55 & 0.68 & 0.45 & 0.62 & \textbf{0.28} & 0.72 & 0.88 & 0.60 \\
    \rowcolor[HTML]{F2F2F2} 
    \textbf{LaRe (Ours)} & \textbf{0.52} & \textbf{0.62} & \textbf{0.39} & \textbf{0.55} & 0.30 & \textbf{0.65} & \textbf{0.78} & \textbf{0.54} \\
    \midrule
    \multicolumn{9}{c}{\textit{Base VLM: Qwen3-VL-4B-Instruct}} \\
    \hline
    Standard CoT & 2.15 & 2.45 & 1.65 & 2.12 & 1.15 & 2.25 & 2.85 & 2.09 \\
    MM-CoT       & 2.08 & 2.32 & 1.52 & 2.25 & 1.12 & 2.18 & 2.65 & 2.02 \\
    CCoT         & 1.75 & 2.38 & 1.72 & 2.45 & 1.28 & 2.65 & 2.72 & 2.14 \\
    ICoT         & 1.92 & 2.25 & 1.55 & 2.18 & 1.22 & 2.35 & 2.82 & 2.04 \\
    CoF          & 2.25 & 2.58 & 1.82 & 2.35 & 1.52 & 3.15 & 3.42 & 2.44 \\
    MCOUT        & 1.05 & 1.25 & 0.68 & 0.85 & 0.58 & 0.95 & 1.38 & 0.96 \\
    LVR          & 1.02 & 1.18 & 0.62 & 0.78 & \textbf{0.52} & 0.88 & 1.32 & 0.90 \\
    \rowcolor[HTML]{F2F2F2} \textbf{LaRe (Ours)} & \textbf{0.98} & \textbf{1.14} & \textbf{0.57} & \textbf{0.73} & \textbf{0.52} & \textbf{0.85} & \textbf{1.25} & \textbf{0.82} \\
    \bottomrule
    \end{tabular}
    \end{adjustbox}
\end{table*}

\paragraph{Analysis of Reasoning Efficiency.}

LaRe exhibits a favorable balance in terms of inference efficiency, as illustrated in Table~\ref{tab:latency}. Although the iterative latent refocusing mechanism introduces additional forward pass overhead at each step, this strategy of substituting generation with latent computation effectively compresses the final textual path. Compared to conventional explicit Chain of Thought approaches, LaRe requires substantially fewer tokens, which directly diminishes the time spent on computationally intensive autoregressive decoding. While the latency advantage remains relatively modest in straightforward perceptual tasks such as POPE due to the inherent fixed iteration costs, the framework justifies its computational allocation in complex reasoning scenarios like MMMU-Pro and $V^*$. In these cases, the model prioritizes processing time for high-quality visual evidence extraction rather than generating redundant textual descriptions. This mechanism enables superior reasoning accuracy while maintaining competitive inference latency.

\begin{figure}[ht]
    \centering
    \includegraphics[width=\linewidth]{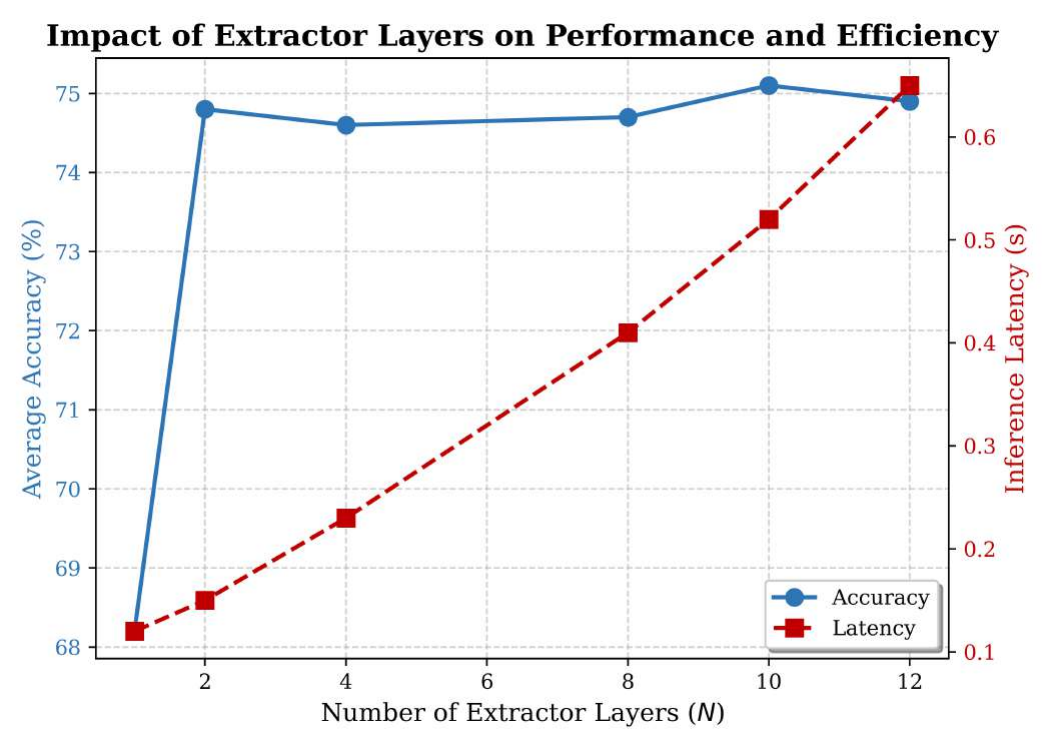}
    \caption{Impact of Extractor depth $N$ on reasoning performance and efficiency. The blue solid line (left y-axis) represents the average accuracy (\%), while the red dashed line (right y-axis) denotes the inference latency per step (seconds). Performance saturates beyond $N=4$, justifying our choice of $N=2$ for a lightweight and efficient implementation.}
    \label{fig:depth_ablation}
\end{figure}

We further analyze the trade-off between reasoning accuracy and computational overhead by adjusting the number of extractor layers $N \in \{1, 2, 4, 8, 10, 12\}$. As illustrated in Figure \ref{fig:depth_ablation}, accuracy improves significantly as $N$ increases from 1 to 2, after which performance gains begin to saturate. Specifically, increasing $N$ beyond 4 yields marginal improvements of less than 0.5\%, whereas inference latency continues to grow. To maintain optimal efficiency and a lightweight resource footprint that supports the claim of reducing token consumption, $N=2$ is selected as the default configuration.

\paragraph{Causal Intervention Analysis.}

To rigorously verify whether the attention weights generated by the extractor reflect the causal necessity of visual evidence or merely constitute superficial pseudo-alignment, we design a causal intervention experiment. During inference, we systematically perturb the visual tokens delivered to the model through four distinct masking protocols. These protocols involve suppressing the top 20\% of visual patches with the highest attention scores, suppressing the bottom 20\% with the lowest scores, applying random occlusion to 20\% of patches across the full image, and completely masking all regions throughout every iteration step. By observing the performance degradation of the model on several challenging benchmarks, we directly quantify the semantic indispensability of the regions prioritized by LaRe, as shown in Table~\ref{tab:intervention}.

\begin{table*}[h]
\centering
\caption{Causal intervention results under various masking strategies using the Qwen3-VL-4B-Instruct backbone, where $\Delta$ denotes the performance drop relative to the full LaRe model without masking.}
\label{tab:intervention}
\begin{adjustbox}{width=\textwidth}
\begin{tabular}{l|cc|cc|cc|cc|cc}
\hline
Masking strategy & MMStar & $\Delta$ & $V^*$ & $\Delta$ & MMVP & $\Delta$ & POPE & $\Delta$ & MMMU-Pro & $\Delta$ \\
\hline
LaRe (Full) & 72.8 & - & 87.0 & - & 57.6 & - & 77.4 & - & 58.4 & - \\

Mask High-Attn & 61.4 & -11.4 & 77.8 & -9.2 & 46.3 & -11.3 & 71.1 & -6.3 & 52.9 & -5.5 \\

Mask Low-Attn & 71.7 & -1.1 & 86.2 & -0.8 & 56.3 & -1.3 & 76.9 & -0.5 & 57.8 & -0.6 \\

Random Mask & 67.5 & -5.3 & 82.9 & -4.1 & 51.8 & -5.8 & 74.3 & -3.1 & 55.7 & -2.7 \\

Mask All Steps & 58.9 & -13.9 & 75.3 & -11.7 & 43.6 & -14.0 & 69.6 & -7.8 & 51.4 & -7.0 \\
\hline
\end{tabular}
\end{adjustbox}
\end{table*}

The experimental results exhibit a pronounced asymmetry across masking strategies, confirming the high semantic fidelity of our latent space refocusing mechanism. Masking the top 20\% of patches with the highest attention scores under strategy A induces substantial performance degradation across all evaluation sets, with accuracy dropping by 11.4\% on MMStar and 11.3\% on MMVP. Conversely, discarding the bottom 20\% of patches with the lowest attention scores under strategy B yields negligible decay, maintaining an average decrease below 1.0\%. This behavior demonstrates that the extractor successfully filters task-irrelevant visual background noise. Moreover, the performance decline observed in strategy A substantially exceeds that of the random masking baseline in strategy C. This disparity verifies that LaRe isolates highly specific and non-trivial visual evidence critical for reasoning, rather than capturing generic saliency structures. Complete removal of the target regions across all iterations under strategy D triggers the most severe cumulative collapse in reasoning accuracy, exemplified by a 14.0\% drop on MMVP. Taken together, these empirical causal interventions demonstrate that the visual regions prioritized by our latent space refocusing mechanism are not merely correlated with textual generation. They constitute a causal necessity for accurate multimodal reasoning.

\begin{table}[H]
    \centering 
    \caption{Probing analysis results for semantic alignment of latent thoughts.}
    \label{tab:probing_analysis}
    \begin{adjustbox}{width=\columnwidth}
    \begin{tabular}{ccccc}
    \toprule
    \multirow{2}{*}{Backbone} & \multirow{2}{*}{Module} & \multirow{2}{*}{Classfier} & \multicolumn{2}{c}{Accuracy} \\
  \cline{4-5}
  &&&Vision &Text\\
    \midrule
    \multirow{4}{*}{Qwen3-VL-2B-Instruct}  
     & w/o Extractor & SVM & 58.4 & 61.4 \\
     & with Extractor & SVM & 94.9 & 95.2 \\
     & w/o Extractor & MLP & 59.6 & 62.2 \\
     & with Extractor & MLP & 96.2 & 96.4 \\
    \hline
    \multirow{4}{*}{Qwen3-VL-4B-Instruct}  
     & w/o Extractor & SVM & 59.2 & 62.5 \\
     & with Extractor & SVM & 95.7 & 95.3 \\
     & w/o Extractor & MLP & 60.0 & 63.5 \\
     & with Extractor & MLP & 94.8 & 95.0 \\
    \bottomrule
    \end{tabular}
\end{adjustbox}
\end{table}

\paragraph{Sequential Dependency Analysis.}
To more directly characterize the sequential dependency among latent tokens, we design a stepwise intervention experiment where at step $i$ we remove the corresponding latent token and measure the downstream task accuracy as well as the output length of the generated responses. The results are summarized in Table~\ref{tab:sequential_dependency}.

\begin{table}[t]
\centering
\caption{Stepwise intervention results of latent tokens. Interventions on earlier steps cause larger accuracy drops, exhibiting a characteristic cascading effect of sequential dependency.}
\label{tab:sequential_dependency}
\begin{adjustbox}{width=\columnwidth}
\begin{tabular}{lcc}
\toprule
Ablated Step & Avg Acc (\%) & Output Tokens \\
\midrule
None (full) & 74.7 & 65.0 \\
Step 1 & 68.4 & 89.3 \\
Step 2 & 70.9 & 78.6 \\
Step 3 & 72.8 & 71.2 \\
Step 4 & 73.6 & 67.4 \\
\midrule
All steps & 61.2 & 102.7 \\
\bottomrule
\end{tabular}
\end{adjustbox}
\end{table}

The experimental results exhibit a clear cascading effect: interventions on earlier steps cause substantially larger accuracy drops than those on later steps (for example, step 1 leads to a 6.3\% drop while step 4 only a 1.1\% drop), accompanied by a corresponding increase in output length. This indicates that the model is forced to compensate for the missing latent information through text. This asymmetric, order-sensitive performance degradation is a hallmark of sequential dependency.

\paragraph{Sensitivity Analysis of Step Allocation.}

To further examine whether a fixed number of steps suffices for queries of varying complexity, we perform a sensitivity analysis of LaRe (4B) on seven benchmarks covering perception, mixed, and reasoning tasks, as shown in Table~\ref{tab:sensitivity_K}.

\begin{table*}[t]
\centering
\caption{Sensitivity analysis of LaRe (4B) with respect to the number of refocusing steps $K$ across seven benchmarks. The fixed $K{=}4$ configuration performs within 0.2\% of the per-benchmark optimum $K^{*}$, regardless of task type or complexity. The best result in each row is shown in \textbf{bold}.}
\label{tab:sensitivity_K}
\setlength{\tabcolsep}{5pt}
\renewcommand{\arraystretch}{1.15}
\begin{tabular}{l c cccccc c c}
\toprule
\multirow{2}{*}{\textbf{Benchmark}} & \multirow{2}{*}{\textbf{Type}} & \multicolumn{6}{c}{\textbf{Number of Refocusing Steps} $K$} & \multirow{2}{*}{$K^{*}$} & \multirow{2}{*}{$\Delta$ vs. $K{=}4$} \\
\cmidrule(lr){3-8}
 & & $1$ & $2$ & $3$ & $4$ & $5$ & $6$ & & \\
\midrule
MMBench    & Mixed   & 79.1 & 80.7 & 81.5 & 82.1 & \textbf{82.2} & 82.1 & 5 & +0.1 \\
MMStar     & Mixed   & 69.2 & 71.0 & 72.1 & \textbf{72.8} & 72.7 & 72.7 & 4 & \phantom{+}0.0 \\
MMVP       & Percep. & 52.0 & 54.4 & 56.3 & \textbf{57.6} & \textbf{57.6} & 57.5 & 4 & \phantom{+}0.0 \\
ScienceQA  & Mixed   & 83.2 & 85.6 & 86.9 & 87.5 & \textbf{87.7} & 87.6 & 5 & +0.2 \\
POPE       & Percep. & 74.3 & 76.0 & 76.9 & \textbf{77.4} & 77.2 & 77.2 & 4 & \phantom{+}0.0 \\
V$^{*}$    & Percep. & 82.4 & 84.7 & 86.2 & \textbf{87.0} & 86.9 & 86.9 & 4 & \phantom{+}0.0 \\
MMMU-Pro   & Reason. & 53.6 & 55.9 & 57.5 & 58.4 & 58.3 & \textbf{58.5} & 6 & +0.1 \\
\bottomrule
\end{tabular}
\end{table*}

Across all seven benchmarks, the fixed configuration $K{=}4$ performs within 0.2\% of the per-benchmark optimum $K^{*}$, irrespective of task type or complexity. This indicates that the potential improvement recoverable by an adaptive strategy is very limited, and further supports the conclusion that using a small fixed number of steps serves as a robust and effective choice for LaRe.

\paragraph{Probing Analysis.}
To examine whether the latent tokens jointly encode linguistic reasoning and visual semantics, we conduct a probing analysis with two weak classifiers including SVM \citep{cortes1995support} and MLP \citep{rumelhart1986learning}. As shown in Table \ref{tab:probing_analysis}, LaRe achieves latent-vision and latent-text matching accuracy close to 95\% across different backbone models. In contrast, removing the Extractor leads to a drop in probing accuracy to approximately 60\%. These results provide strong evidence for the critical role of the Extractor, demonstrating its effectiveness in structuring cross-modal inputs into unified semantic representations within the latent space.

\begin{figure*}[ht]
    \centering
    \includegraphics[width=0.7\linewidth]{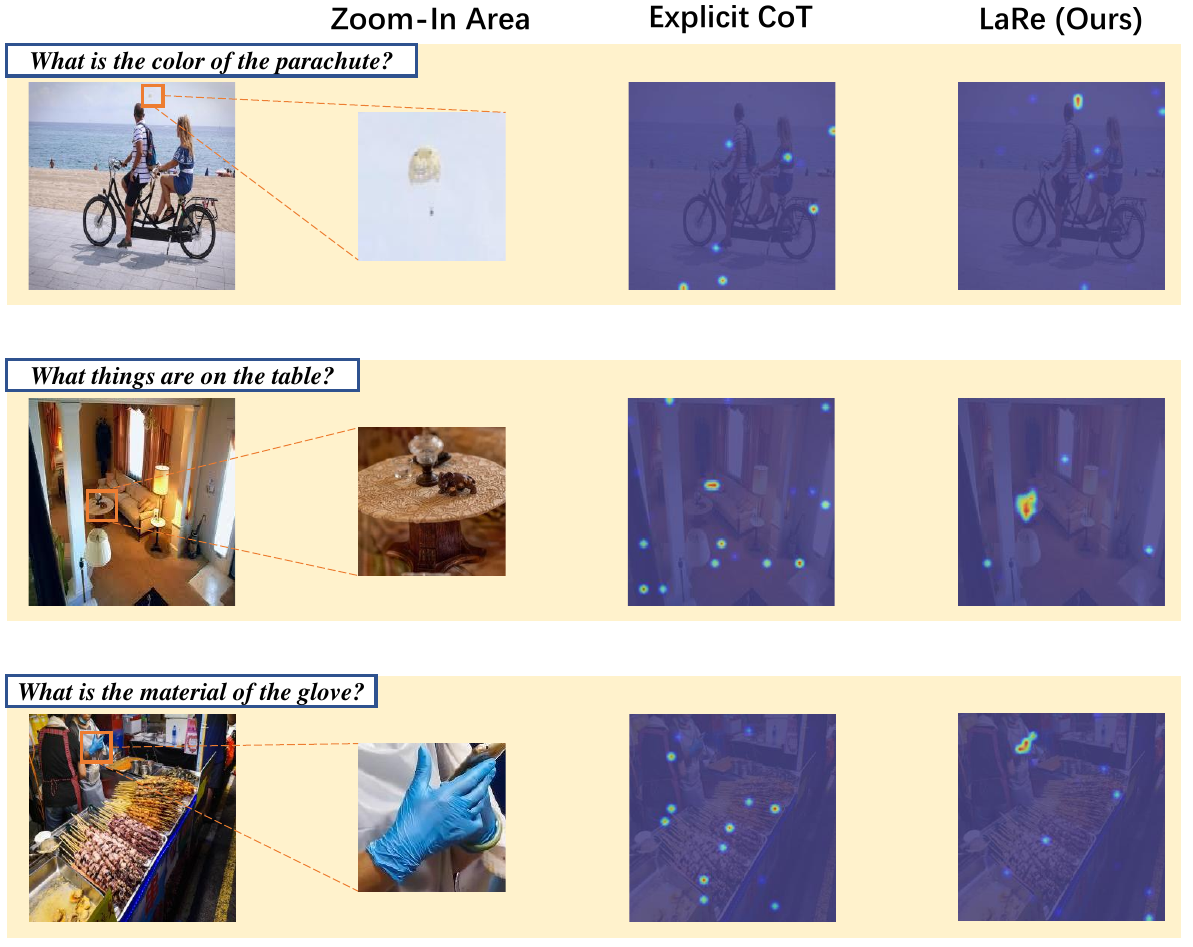}
    \caption{Comparison of attention heatmaps between explicit CoT and LaRe. Unlike conventional explicit CoT which is easily distracted by prominent or complex backgrounds, LaRe utilizes a latent refocusing mechanism to accurately isolate small or occluded reasoning evidence.}
    \label{fig:case_study}
\end{figure*}

\subsection{Generalization across VLM Architectures}
To further verify that the effectiveness of LaRe does not depend on a specific VLM architecture, we conduct supplementary experiments on two additional backbone models with markedly different designs: LLaVA-1.5-7B~\citep{liu2023visualinstructiontuning} and InternVL2.5-8B~\citep{chen2025expandingperformanceboundariesopensource}. For each backbone, under identical training and inference settings, we compare LaRe against the standard CoT baseline and the latent-space reasoning method LVR.

\begin{table*}[t]
\centering
\caption{Generalization of LaRe on Additional VLM Backbones. We report accuracy on seven multi-modal benchmarks and the average number of reasoning tokens (\#Tok). The best result within each backbone group is shown in \textbf{bold}.}
\label{tab:generalization}
\setlength{\tabcolsep}{4pt}
\renewcommand{\arraystretch}{1.15}
\resizebox{\textwidth}{!}{%
\begin{tabular}{llccccccccc}
\toprule
Base VLM & Method & MMBench & ScienceQA & MMVP & V$^{*}$ & MMStar & MMMU-Pro & POPE & Avg. & \#Tok \\
\midrule
\multirow{3}{*}{LLaVA-1.5-7B}
 & CoT  & 66.4 & 69.8 & 22.0 & 47.3 & 36.2 & 16.8 & 85.4 & 49.1 & 512 \\
 & LVR  & 67.8 & 71.2 & 27.6 & 51.8 & 38.1 & 18.2 & 86.1 & 51.5 & 487 \\
 & \textbf{LaRe} & \textbf{70.1} & \textbf{73.5} & \textbf{34.7} & \textbf{57.8} & \textbf{41.5} & \textbf{22.3} & \textbf{87.2} & \textbf{55.3} & \textbf{196} \\
\midrule
\multirow{3}{*}{InternVL2.5-8B}
 & CoT  & 79.6 & 95.4 & 42.7 & 63.5 & 60.1 & 48.2 & 87.6 & 68.2 & 486 \\
 & LVR  & 80.9 & 95.9 & 47.2 & 67.4 & 62.0 & 50.1 & 88.2 & 70.2 & 463 \\
 & \textbf{LaRe} & \textbf{83.2} & \textbf{96.8} & \textbf{55.3} & \textbf{73.8} & \textbf{65.4} & \textbf{54.7} & \textbf{89.3} & \textbf{74.1} & \textbf{178} \\
\bottomrule
\end{tabular}}
\end{table*}

As shown in Table~\ref{tab:generalization}, on both additional architectures, LaRe consistently outperforms the two baseline methods across all seven benchmarks. Specifically, compared to standard CoT, LaRe improves the average accuracy by 6.2\% on LLaVA-1.5-7B and by 5.9\% on InternVL2.5-8B, while reducing the number of reasoning tokens by over 60\%. This performance gain is highly consistent with the results observed on the Qwen3-VL backbone, indicating that the proposed latent refocusing paradigm and semantic augmentation training strategy generalize well across VLMs that differ in visual encoders, projection modules, and pre-training schemes.

\section{Case Study}
In this section, we present a qualitative attention comparison across three challenging perception-intensive tasks to empirically demonstrate the advantages of LaRe, as illustrated in Figure~\ref{fig:case_study}.

In the first case concerning long-range retrieval of a parachute color, the target occupies a minimal pixel footprint. The standard explicit CoT fails to localize the object during discrete token generation, causing attention to scatter across the foreground bicycle, the rider, and salient ground noise. In contrast, LaRe utilizes a lightweight semantic aggregation module driven by the latent decision states of the large language model. This Extractor enables dynamic visual focusing within the latent space, which concentrates high attention weights precisely on the small parachute pixels in the sky.

A similar pattern emerges in the second case, which evaluates item identification on a table within a cluttered indoor scene. The explicit CoT becomes distracted by large redundant background regions such as sofas, leading to substantial visual deviation. LaRe suppresses this background interference through a latent refocusing mechanism, directing attention exclusively to the round wooden table in the corner and its small items.

The third case assesses local attribute recognition for glove material. The standard explicit CoT becomes anchored to the dense and highly salient barbecue skewers in the foreground, causing global attention to disperse across the food items without extracting the core answer. LaRe filters out low-level visual saliency interference and anchors attention precisely on the blue plastic gloves worn by the staff member at the top of the image.

These observations confirm that latent space reasoning effectively mitigates the limitations of explicit sequential generation in perception-intensive scenarios.

\section{Discussion}
The proposed LaRe paradigm achieves dual advantages in token efficiency and reasoning performance for multimodal tasks through an iterative latent refocusing mechanism and a diffusion model based semantic enhancement training strategy. However, as noted in the Limitations section, the current framework still leaves several issues worthy of deeper investigation in future work, and these issues also point to potential directions for further improvement.

The current LaRe framework relies on a preset, static number of latent iterations $K$. Although we have empirically determined an optimal threshold at which performance saturates, this one size fits all strategy struggles to accommodate the substantial variability in task complexity across different multimodal benchmarks. For simple visual question answering tasks, a fixed number of iterations may introduce redundant computational overhead, whereas for complex tasks involving multi step spatial reasoning or fine grained perception, a preset upper bound may restrict the model's reasoning potential. Future work can explore a dynamic halting mechanism, for instance by drawing on ideas from Adaptive Computation Time (ACT)~\citep{graves2016adaptive} or PonderNet~\citep{banino2021pondernet}, allowing the model to adaptively decide when to terminate iterative reasoning based on the convergence of latent states, the confidence of hidden representations, or the trend of reconstruction loss. 

Going a step further, we believe that a learnable stopping policy based on reinforcement learning is a particularly promising direction to explore. The iterative refocusing process of LaRe is inherently a sequential decision problem, where each iteration corresponds to a decision of whether to continue gathering more visual evidence. This aligns well with the Markov Decision Process (MDP) formulation of reinforcement learning. One can treat the hidden state at each step as a state, the continue or stop choice as a discrete action, and construct a composite reward function using the two natural supervision signals available in the LaRe framework, namely task accuracy and diffusion reconstruction loss. This would simultaneously constrain reasoning performance, computational efficiency, and latent space quality. Coupled with recent successful applications of policy optimization methods such as GRPO~\citep{shao2024deepseekmath} and PPO~\citep{schulman2017proximal} to control the reasoning length of large language models, this think on demand paradigm not only promises to further improve LaRe's ability to balance efficiency and flexibility, but also may give rise to a multimodal reasoning framework with metacognitive capabilities. That is, the model would not only produce an answer, but also autonomously determine how much thinking is sufficient to arrive at a reliable answer.

\section{Usage of LLMs}
In this paper, LLMs were used for coding assistance and writing support. Specifically, they were employed to generate code snippets, debug existing scripts, and optimize algorithms, significantly accelerating the development cycle. For writing support, they were primarily utilized for proofreading the text and formatting \textit{LaTeX} code.

\end{document}